%% file: MultiLevelDIF.tex
\documentclass[10pt,twocolumn,letterpaper]{article}

\usepackage{iccv}
\usepackage{titling} %
\usepackage{times}
\usepackage{epsfig}
\usepackage{graphicx}
\usepackage{amsmath}
\usepackage{amssymb}
\usepackage{subcaption}
\usepackage{booktabs,tabularx,tabulary}
\usepackage{multirow}
\usepackage{float}
\usepackage{stfloats}
\usepackage[accsupp]{axessibility}

\usepackage[breaklinks=true,colorlinks,bookmarks=false]{hyperref}

\iccvfinalcopy %

\def\iccvPaperID{6006} %

\ificcvfinal\fi

\input{0_tweaks.tex}

\input{0_math.tex}

\predate{}
\date{}
\postdate{}

\begin{document}

\title{Multiresolution Deep Implicit Functions for 3D Shape Representation}

\author{Zhang Chen$^{1,2,}$\thanks{Work done while the author was an intern at Google.}
\qquad
Yinda Zhang$^{1}$
\qquad
Kyle Genova$^{1}$
\qquad
Sean Fanello$^{1}$
\qquad
Sofien Bouaziz$^{1}$\\
Christian H{\"a}ne$^{1}$
\qquad
Ruofei Du$^{1}$
\qquad
Cem Keskin$^{1}$
\qquad
Thomas Funkhouser$^{1}$
\qquad
Danhang Tang$^{1}$\\
$^{1}$ Google \qquad $^{2}$ ShanghaiTech University
}

\maketitle
\ificcvfinal\thispagestyle{empty}\fi

\input{0_abstract.tex}
\input{1_intro.tex}

\input{2_related.tex}

\input{3_background.tex}
\input{4_method.tex}

\input{5_exp.tex}
\input{6_conclusion.tex}

\clearpage
\title{\Large \bf Multiresolution Deep Implicit Functions for 3D Shape Representation \\ (Supplementary Material)}
\maketitle
\input{7_supp.tex}
\clearpage

{\small
\bibliographystyle{ieee_fullname}
\bibliography{MultiLevelDIF}
}

\end{document}

%% file: 0_tweaks.tex
\usepackage{overpic}
\usepackage{amssymb}
\usepackage{amsmath}
\usepackage{graphicx}
\usepackage{enumitem}
\usepackage{microtype} %

\usepackage{color}
\definecolor{turquoise}{cmyk}{0.65,0,0.1,0.3}
\definecolor{purple}{rgb}{0.65,0,0.65}
\definecolor{dark_green}{rgb}{0, 0.5, 0}
\definecolor{green}{rgb}{0, 1.0, 0}
\definecolor{orange}{rgb}{0.8, 0.6, 0.2}
\definecolor{red}{rgb}{0.8, 0.2, 0.2}
\definecolor{blueish}{rgb}{0.0, 0.7, 1}
\definecolor{light_gray}{rgb}{0.7, 0.7, .7}
\definecolor{pink}{rgb}{1, 0, 1}

\newcommand{\hide}[1]{{}} %

\newcommand{\Fig}[1]{\autoref{fig:#1}}
\newcommand{\Figure}[1]{\autoref{fig:#1}}
\newcommand{\Table}[1]{\autoref{tab:#1}}

\newcommand{\Eq}[1]{Equation~\ref{eq:#1}}

\newcommand{\Sec}[1]{Section~\ref{sec:#1}}

\usepackage{currfile}

\newcommand{\CIRCLE}[1]{\raisebox{.5pt}{\footnotesize \textcircled{\raisebox{-.6pt}{#1}}}}
\usepackage{amsmath}

\usepackage{lipsum}

\renewcommand{\paragraph}[1]{{\vspace{.25em}\noindent \textbf{#1.}}}

%% file: 0_math.tex
\newcommand{\sdf}{S}
\newcommand{\SDF}{\sdf}
\newcommand{\volume}{V}
\newcommand{\sdfthreshold}{\tau}
\newcommand{\feature}{F}
\newcommand{\encoder}{\mathcal{E}}
\newcommand{\decoder}{\mathcal{D}}
\newcommand{\pointset}{\mathcal{P}}
\newcommand{\nearsurfacesamples}{\pointset_{S}}
\newcommand{\uniformsamples}{\pointset_{U}}
\newcommand{\visiblesamples}{\pointset_{V}}
\newcommand{\occludedsamples}{\pointset_{O}}
\newcommand{\MDIF}{MDIF~}
\newcommand{\loss}{\mathcal{L}}
\newcommand{\levels}{N}
\newcommand{\level}{n}
\newcommand{\Gaussian}{G}
\newcommand{\latent}{\mathbf{z}}
\newcommand{\latentgrid}{Z}
\newcommand{\residual}{R}

\newcommand{\point}{\mathbf{x}}
\newcommand{\gt}[1]{\bar{#1}}
\newcommand{\consistencyweight}{\lambda}
\newcommand{\distancefunction}{d}

%% file: 0_abstract.tex
\begin{abstract}
We introduce Multiresolution Deep Implicit Functions (MDIF), a hierarchical representation that can recover fine geometry detail, while being able to perform global operations such as shape completion. 
Our model represents a complex 3D shape with a hierarchy of latent grids, which can be decoded into different levels of detail and also achieve better accuracy.
For shape completion, we propose latent grid dropout to simulate partial data in the latent space and therefore defer the completing functionality to the decoder side. 
This along with our multires design significantly improves the shape completion quality under decoder-only latent optimization.
To the best of our knowledge, MDIF is the first deep implicit function model 
that can at the same time (1) represent different levels of detail and allow progressive decoding; (2) support both encoder-decoder inference and decoder-only latent optimization, and fulfill multiple applications; (3) perform detailed decoder-only shape completion. Experiments demonstrate its superior performance against prior art in various 3D reconstruction tasks.
\end{abstract}

%% file: 1_intro.tex
\section{Introduction}
In recent years, deep implicit functions (DIF) have gained much popularity as a 3D shape representation in applications such as compression \cite{deepimplicitcompression}, shape completion \cite{dai2017shape}, neural rendering \cite{mildenhall2020nerf,tewari2020state}, and super-resolution~\cite{chibane2020implicit}.
In contrast to explicit representations such as point clouds, voxels, or meshes, a 3D shape is encoded into a compact latent vector, which when combined with a sampled 3D location as input to a decoder can be used to evaluate an implicit function for surface reconstruction. 

In this paper, our objective is to design a DIF for shape representation that has three main properties: \CIRCLE{1} represent shapes with arbitrarily fine details (adding more bits to the representation provides more details), 
\CIRCLE{2} support both encoder-decoder inference and decoder-only latent optimization, and can be applied to different tasks,
and \CIRCLE{3} enable detail-preserving shape completion from inputs with large unobserved regions.
These properties are all important for a shape representation. Yet, to the best of our knowledge, no prior method has achieved all three properties.

Existing DIF methods can be classified into global and local approaches.
Early methods mostly belong to the global category~\cite{park2019deepsdf, chen2019learning, mescheder2019occupancy, xu2019disn, michalkiewicz2019deep}, where a single latent vector is used to represent the whole shape.
These approaches learn to encode a global shape prior in a compact latent space, which can then be leveraged to fulfill various reconstruction tasks.
However, due to the limited capacity of the latent space and the global nature of these approaches, global methods usually lack fine-grained detail. %

\begin{figure}
\center
\includegraphics[trim={0cm 1.5cm 10.5cm 0cm},clip,width=0.9\linewidth]{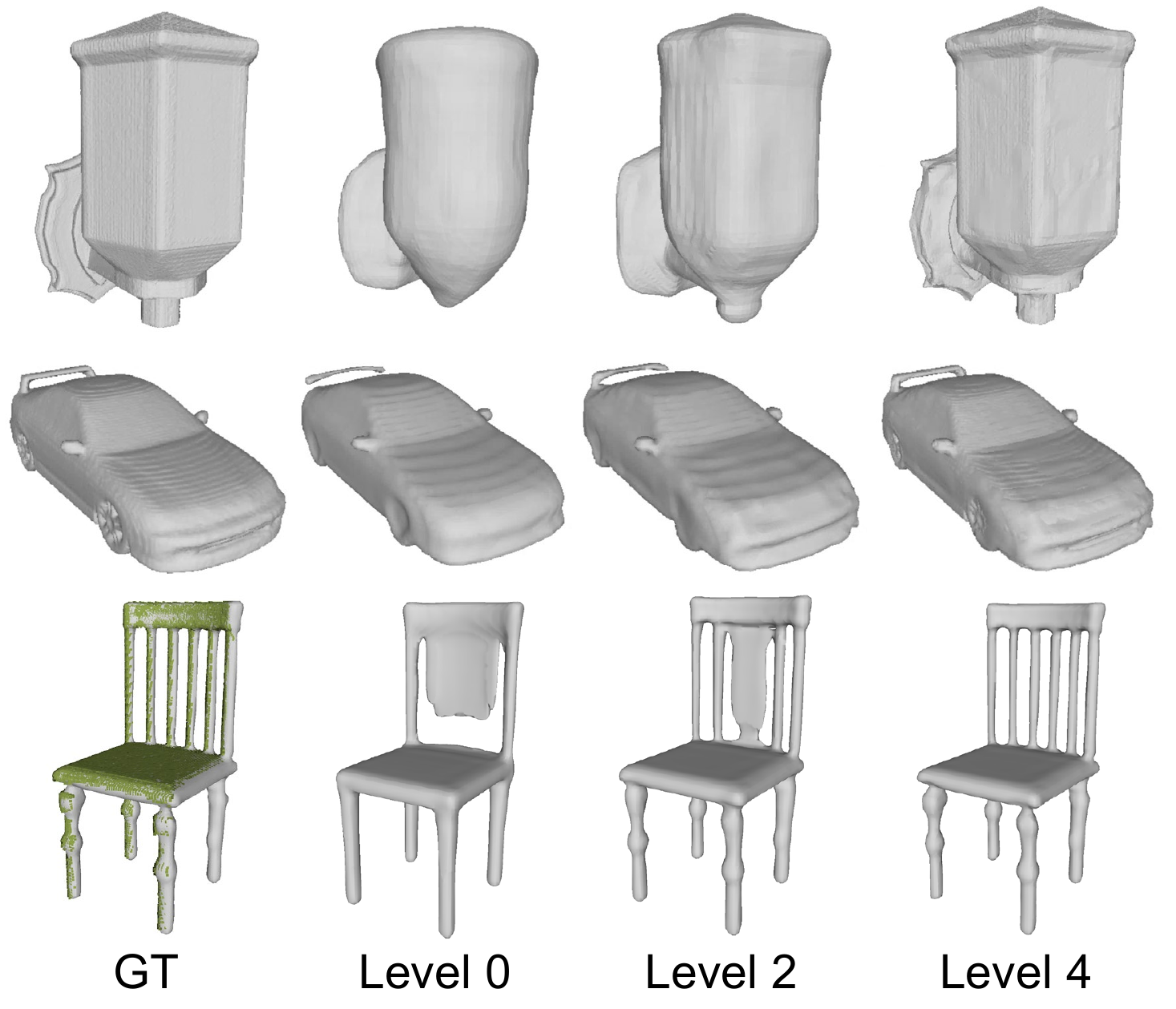}
\vspace{-0.75em}
\caption{Example results of our model for auto-encoding (row 1 and 2) and shape completion (row 3) in different levels of detail. Green dots represent the observed depth pixels for the completion task.}
\label{fig:teaser}
\vspace{-1.75em}
\end{figure}

More recently, local approaches~\cite{jiang2020local, chabra2020deep} have been proposed. These methods divide the space into local regions and encode each one with a latent vector.
Such local representations provide better accuracy and generalization when representing shapes, especially under decoder-only latent optimization. However, they do not model a global prior. As a result, they cannot be used for shape completion with large unobserved regions since in such regions there is no data to optimize the latent vectors.
To overcome this issue, ~\cite{Genova_2020_CVPR, chibane2020implicit} use an encoder to regress local latent vectors from incomplete inputs. However, their methods are limited to encoder-decoder inference when doing shape completion. Compared to decoder-only latent optimization, encoder-decoder inference has less flexibility on the inputs and is less accurate for preserving detail in observed regions.

In this paper, we propose a novel 3D representation: Multiresolution Deep Implicit Function (MDIF). The core idea is to represent a shape as a multiresolution hierarchy of latent vectors, where each level encodes different frequencies of an implicit function. The higher levels of our representation provide the global shape and the lower levels provide fine detail. 
Different from local methods~\cite{Genova_2020_CVPR, chibane2020implicit}, MDIF has a \textit{one-decoder-per-level} architecture, where each decoder produces a residual with respect to its parent level, like a Haar wavelet \cite{chui2016introduction}. This simplifies learning of fine detail and enables progressive decoding to achieve arbitrary levels of detail (see \autoref{fig:teaser}).

To enable detailed shape completion with decoder-only latent optimization, we further propose to use global connection across levels as well as applying dropout on the latent codes. The global connection serves to integrate global priors into lower levels to compensate for missing observations. Meanwhile, applying dropout on the latent codes simulates partial observation in the latent space during training, and therefore forces the decoders to learn to complete shapes under encoder-less scenario.

Overall, our model has the following merits:
\begin{enumerate}
    \item Can represent complex shapes with high accuracy, and allows progressive decoding for different levels of detail.
    
    \item Supports both encoder-decoder inference and decoder-only latent optimization, and is effective for different applications as illustrated in the experimental results.
    
    \item Enables detailed decoder-only shape completion that accurately preserves detail in observed regions while producing plausible results in unobserved regions.
\end{enumerate}

%% file: 2_related.tex
\section{Related Work}

There are largely two types of 3D geometry representations in computer graphics and vision. \textit{Explicit representations} such as meshes, splines, and point clouds, are widely adopted in the field of CAD and animation, since they are compact and highly optimized for editing and rendering.
\textit{Implicit representations}, such as the zero-level set of a signed distance field, have gained increasing popularity in volumetric capture \cite{motion2fusion, kinfu, dynfu, dou2016fusion4d,tsdf}, since they can represent arbitrary surface topology and define watertight surfaces.

Convolutional neural networks (CNNs) have been proposed for predicting an implicit representation of objects. Early techniques were only able to predict low-resolution grids \cite{girdhar2016learning, choy20163d,wu20153d}. More recently, methods relying on an octree structure have been proposed \cite{tatarchenko2017octree,hane2019hierarchical,riegler2017octnetfusion,Wang:2017:OOC:3072959.3073608,wang2020deep} to avoid the cubic growth inherent to high-resolution grids. However, the implicit representation learnt by these networks is still discrete, potentially creating discretization artefacts when reconstructing 3D shapes. To overcome this limitation and allow for learning the implicit representation over the continuous domain, the problem can be reformulated as a multi-layer perceptron (MLP) which takes the location at which the implicit representation is to be evaluated as input \cite{park2019deepsdf, chen2019learning, mescheder2019occupancy, xu2019disn, michalkiewicz2019deep}. This allows for querying the implicit representation at continuous locations during test time.
Termed as \textit{Deep Implicit Functions} (DIF), this technique 
can be categorized into global, local, and hierarchical methods.

\paragraph{Global methods}
Global methods represent a 3D shape with a single holistic latent code. The projection to the latent space can be done via an encoder~\cite{chen2019learning}, or latent optimization~\cite{park2019deepsdf}. A decoder is then used to recover the shape from the latent vector. To obtain a smooth manifold on the latent space for shape generation,
people have developed optimization strategies based on auto-decoding~\cite{park2019deepsdf},
curriculum learning \cite{duan2020curriculum}, and adversarial training~\cite{kleineberg2020adversarial}. Global methods are robust to local noise, hence have good shape completion capability. However, these approaches have difficulty recovering fine detail. 
Recent methods \cite{sitzmann2020implicit,tancik2020fourfeat,mildenhall2020nerf} propose to use periodic activation functions to lift the input positional vector to high dimensional space allowing to better preserve high frequency detail. 
However, these methods focus on per-instance fitting instead of generalization to new scenes and objects.

\paragraph{Local methods} In contrast, local methods uniformly divide the 3D space into local grids~\cite{jiang2020local, chabra2020deep, chibane2020implicit} or use an encoder to decompose space into local parts~\cite{genova2019learning, Genova_2020_CVPR}. Then they either assign each local grid/part with a latent code~\cite{jiang2020local, chabra2020deep, Genova_2020_CVPR} or trilinearly interpolate feature grids to obtain the latent code at each querying location~\cite{chibane2020implicit}.
Since each latent code only needs to represent the shape in a local region, it is much easier to encode detail and generalize to unseen objects. However, these methods do not include global context, hence it is not feasible to perform decoder-only shape completion when there are large unobserved regions. 
While the feature grids used by~\cite{chibane2020implicit} span multiple resolutions, they still do not contain global context and are only used to represent single level of detail.

\paragraph{Hierarchical methods} 
Some methods perform shape reconstruction in stages, where a low-resolution shape prediction precedes a high-resolution prediction \cite{dai2017shape,jiang2020sdfdiff,hanocka2019meshcnn,wang2018global}.
For example, Global-To-Local Generative Models~\cite{wang2018global} decode a global voxel grid and then add detail with a part-wise refiner.
NSVF~\cite{liu2020neural} uses an octree hierarchy of implicit functions to represent the radiance field for neural rendering.
Though their motivation for using an octree is similar as ours, they do not use it to represent a globally consistent shape, 
but rather a view-dependent radiance function suitable for view synthesis.
They would not be able, for example, to perform shape completion.

%% file: 4_method.tex
\section{Methodology}

\begin{figure*}[t!]
\centering
\includegraphics[trim={0cm 3.5cm 0cm 0cm},clip,width=0.8\textwidth]{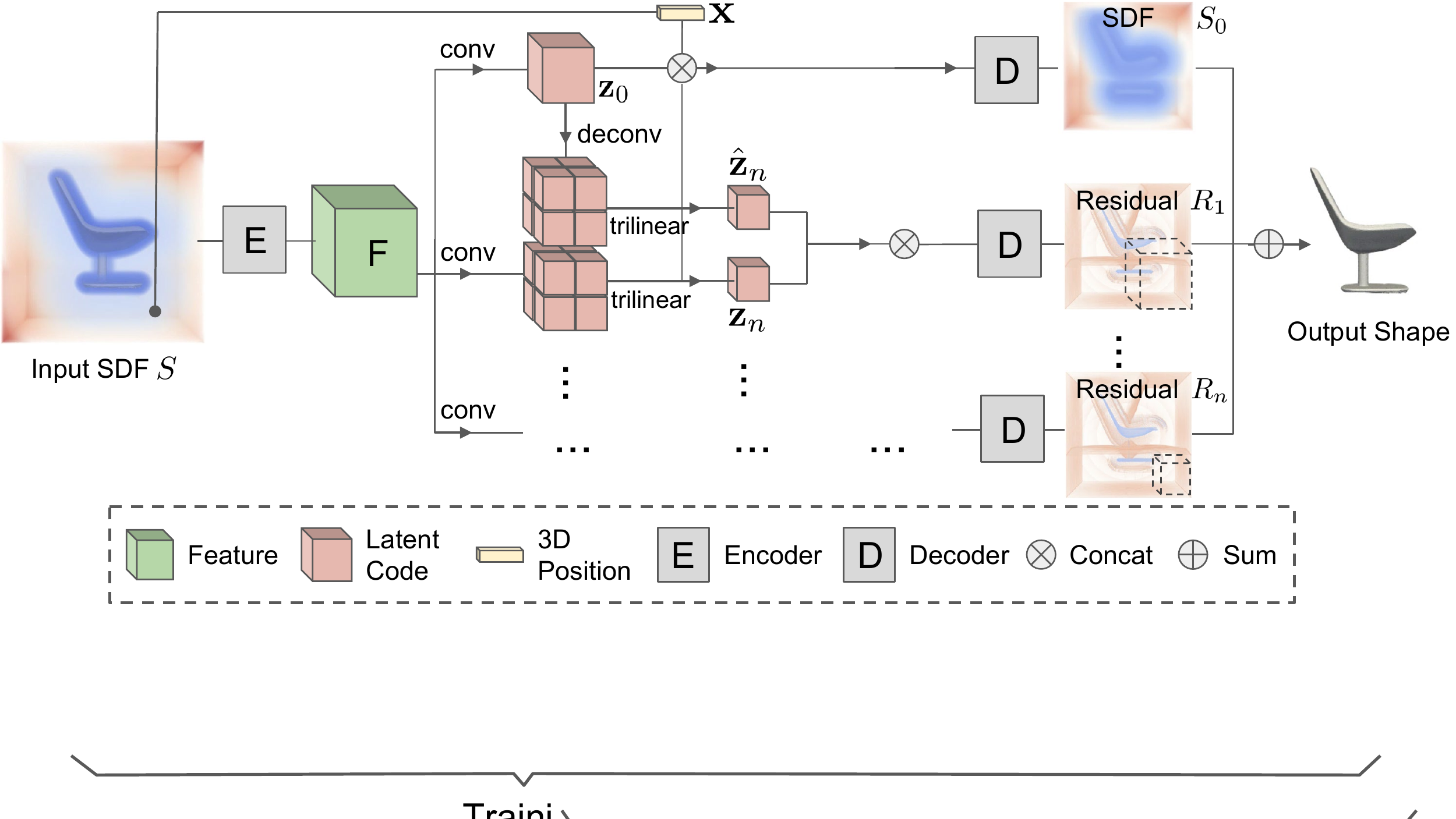}
\vspace{-0.75em}
\caption{Starting from the input SDF $\SDF$, we first extract a global feature $\feature$, which is then encoded into different levels of latent codes through 3D convolutions and transposed convolutions. The decoder is performed per-level to support different resolutions. The outputs of the pipeline consist of a global SDF $\SDF_0$ and multiple residuals $\residual_{\level}$ at different scales, which are used to compute the final reconstruction.}
\vspace{-1.5em}
\label{fig:pipeline}
\end{figure*}

Our overarching goal is to design a flexible representation that can generate shapes from coarse to fine resolutions for reconstruction or completion tasks. Depending on the application, our model can perform inference in the encoder-decoder mode for efficiency or the decoder-only latent optimization for better accuracy. To achieve this, our pipeline, shown in~\Figure{pipeline}, encodes the input SDF into multiple levels of latent codes. Each level has a decoder reconstructing in a different detail level. To detail our design, we first formulate a multi-resolution representation in the form of traditional implicit function in~\Sec{decomposition}. Then in~\Sec{architecture}, we explain how to design a deep neural network version of this representation. The training process is described in~\Sec{training}. Finally in~\Sec{inference}, we explain different inference modes with respect to different applications. 

\subsection{Multires Implicit Function}
\label{sec:decomposition}
We choose to learn the signed distance function (SDF), which is a level set defined as:
\begin{equation}
    \volume(\sdfthreshold) = \{\point: \SDF(\point) = \sdfthreshold\}
\end{equation}
where $\volume$ is the volume containing the shape, $\point$ is a 3D point inside $\volume$, and $\SDF:\mathbb{R}^3 \rightarrow \mathbb{R}$ is the SDF function that represents the signed distance to the closest surface (positive on the outside and negative on the inside). We then use $\SDF(\point)$ to represent the SDF value of a particular point $\point$, and $\volume(0)$ to represent the surface or zero-crossing.

Now we can define an $\levels$-level version of $\SDF$ as $\{S_{\level}\}$,  $\level=0\dots\levels-1$, where each level represents different frequency of details from low to high. To construct this, we subdivide $\volume$ into an $\levels$-level octree. Unlike conventional octrees which only subdivide non-empty cells, our tree is balanced because completing partial observation is one of our target scenarios. 

\begin{figure}[t]
\centering
\includegraphics[trim={0cm 7cm 13.5cm 0cm},clip,width=0.6\linewidth]{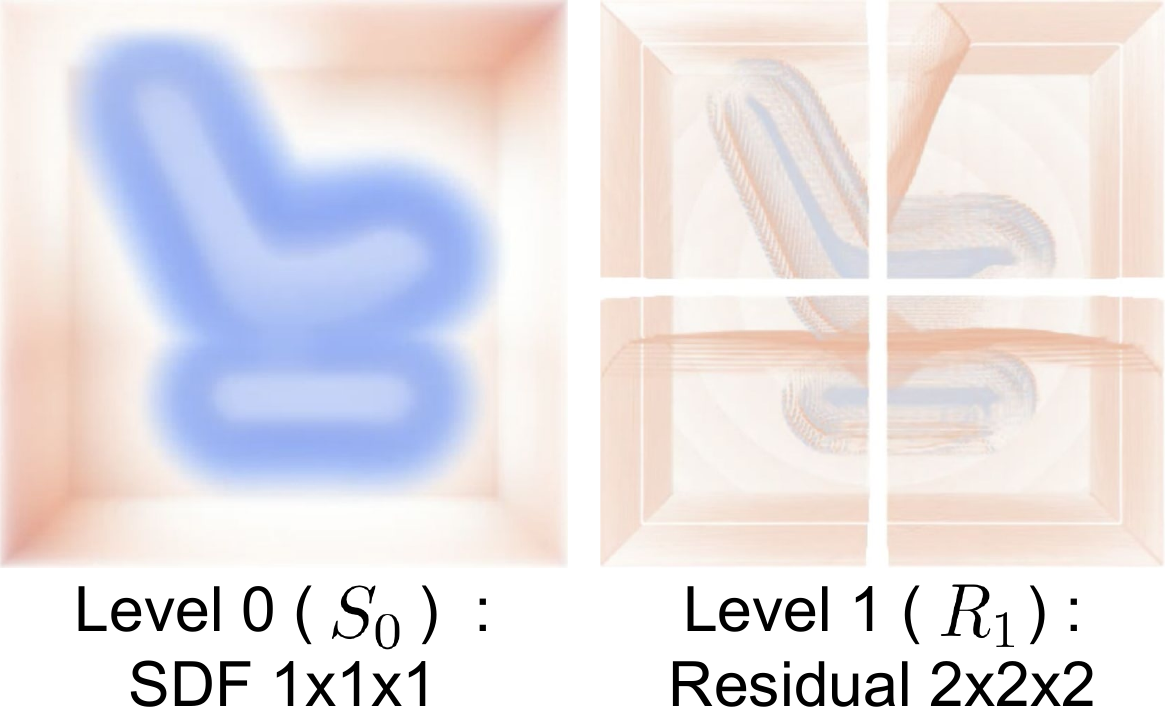}
\vspace{-0.5em}
\caption{Octree subdivision and decoded outputs of the first two levels: (left) level 0 contains a single cell and decodes into SDF; (right) level 1 contains a $2^3$ grid and decodes into residuals. Aggregating all levels we have the final SDF with fine details.}
\vspace{-1.75em}
\label{fig:octree}
\end{figure}

For level 0~(the coarsest level), geometry is represented as SDF $\SDF_{0}$; for level $\level > 0$, we use the residual $\residual_{\level}=\SDF_{\level}-\SDF_{\level-1}$ to capture finer details, as shown in~\Fig{octree}. The final SDF reconstruction is therefore defined as
$\SDF = \SDF_{0} + \sum_{\level=1}^{\levels-1} \residual_{\level}$.
In \Sec{ablation}, we empirically show that inferring residuals yields better performance compared to directly regressing the SDF.

\subsection{Multires Deep Implicit Function}
\label{sec:architecture}
The idea of a \textit{deep} version of the multires implicit function, is to encode the shape in each cell of the octree into a latent code $\latent$ with DNN. For a cell in level $\level=0$, its latent code represents an SDF; while for a cell in $\level>0$, its latent code encodes residuals. Eventually we end up having a tree of latent codes $\latentgrid$, where the latent codes in each cell of level $\level$ form a latent grid $\latentgrid_n$ at this level. The spatial resolution and total capacity of the latent grids increase with the level, and consequently the level of detail gets higher.

In~\Figure{pipeline}, we describe the design of our network architecture to encode the shape into $\latentgrid$.
On the encoder side, the input is the regular grid form of a SDF $\SDF$ with a resolution of $128^3$. The encoder $\encoder$ first extracts a global feature $\feature$ from $\SDF$. 
Then $\feature$ is encoded into different levels of latent grids through 3D convolution layers. %
Note that at level 0, there is only one latent code representing the global shape which is critical for completion tasks.

On the decoder side, unlike~\cite{chibane2020implicit}, our model has one decoder per level to support different levels of detail. For the decoder module at each level, we choose IM-Net~\cite{chen2019learning} which consists of several fully-connected layers. 
The remaining question is \emph{what do we input to the decoders?} 
At the global level ($\level=0$), since there is only one latent code, the decoder $\decoder_{0}$ simply takes $\latent_{0}$ and a 3D position $\point$ as input, and decodes the SDF value at that point. For higher levels ($\level>0$), the input of decoder $\decoder_{\level}$ consists of two parts. The first part is similar to~\cite{chibane2020implicit}, we use trilinear interpolation to sample a latent code $\latent_{n}$ from the latent grid of this level as $\latentgrid_n(\point)$, based on the 3D location $\point$. For the second part, we first apply deconvolution to upsample $\latent_{0}$ to a latent grid $\hat{\latentgrid}_n$, which has the same spatial resolution as $\latentgrid_n$. Then trilinear interpolation is also applied to sample a latent code $\hat{\latent}_n$ from $\hat{\latentgrid}_n$. This allows the decoder $\decoder_{n}$ to have access to the global context to better decode local details as well as compensating for missing data during shape completion. We call this \textit{global connection}. Formally,
\begin{equation} 
\begin{split}
\decoder_{0}(\latent_{0}, \point) &= \SDF_0,\\ 
\decoder_{\level}(\latent_{\level}, \hat{\latent}_{\level}) &= \residual_{\level}, \level>0.
\end{split}
\label{eq:residual}
\end{equation}
Note that for $\level>0$, the decoders do not need to take 3D positions $\point$ as input, because $\latent_n$ and $\hat{\latent}_n$ are already functions of $\point$ via trilinear interpolation. Finally, since $\decoder_{n>0}$ predicts residual $\residual$, the outputs of all levels are aggregated to have the final SDF. For detailed network architecture, please refer to our supplementary.

\subsection{Training}
\label{sec:training}
\MDIF is trained end-to-end in an encoder-decoder fashion because: \CIRCLE{1} it allows both encoder-decoder inference and decoder-only latent optimization to be available during test-time; \CIRCLE{2} training with an encoder is generally more efficient comparing to training in decoder-only mode, since latent codes are not initialized randomly.

\paragraph{Points Sampling} We generate $128^3$ regular SDF grids as the input of the encoder $\encoder$. In addition, the decoders require a 3D point set as training data. 
Similar to~\cite{Genova_2020_CVPR}, we sample a uniform point set $\uniformsamples$ inside the object bounding box, as well as a near-surface point set $\nearsurfacesamples \subset \{(\point, \SDF(\point)): |\SDF(\point)| < 0.04 \}$ for each training object. Each point set has 100K samples. Mixing the two gives us the final training set $\pointset = \uniformsamples \cup \nearsurfacesamples $, which implicitly applies more weight to the near-surface points. At each training iteration, 4096 samples are randomly drawn from each set.

\paragraph{Loss} During training, our final loss is the summation of losses at all levels, such that $\loss = \sum_{\level=0}^\levels\loss_{\level}$. For each level $\level$, we first aggregate the predicted SDF and residual up to this level to produce $\SDF_{\level}$, and then measures the L1 difference between it and groundtruth $\gt{\SDF}$. Formally,
\begin{equation} 
\loss_{n} = \frac{1}{|\pointset|}\sum_{\point \in  \pointset} \left|\SDF_n(\point) - \gt{\SDF}(\point)\right|.\\ 
\label{eq:loss_train}
\end{equation}

\begin{figure}[t]
\centering
\includegraphics[trim={0cm 5.2cm 7.5cm 0cm},clip,width=0.9\linewidth]{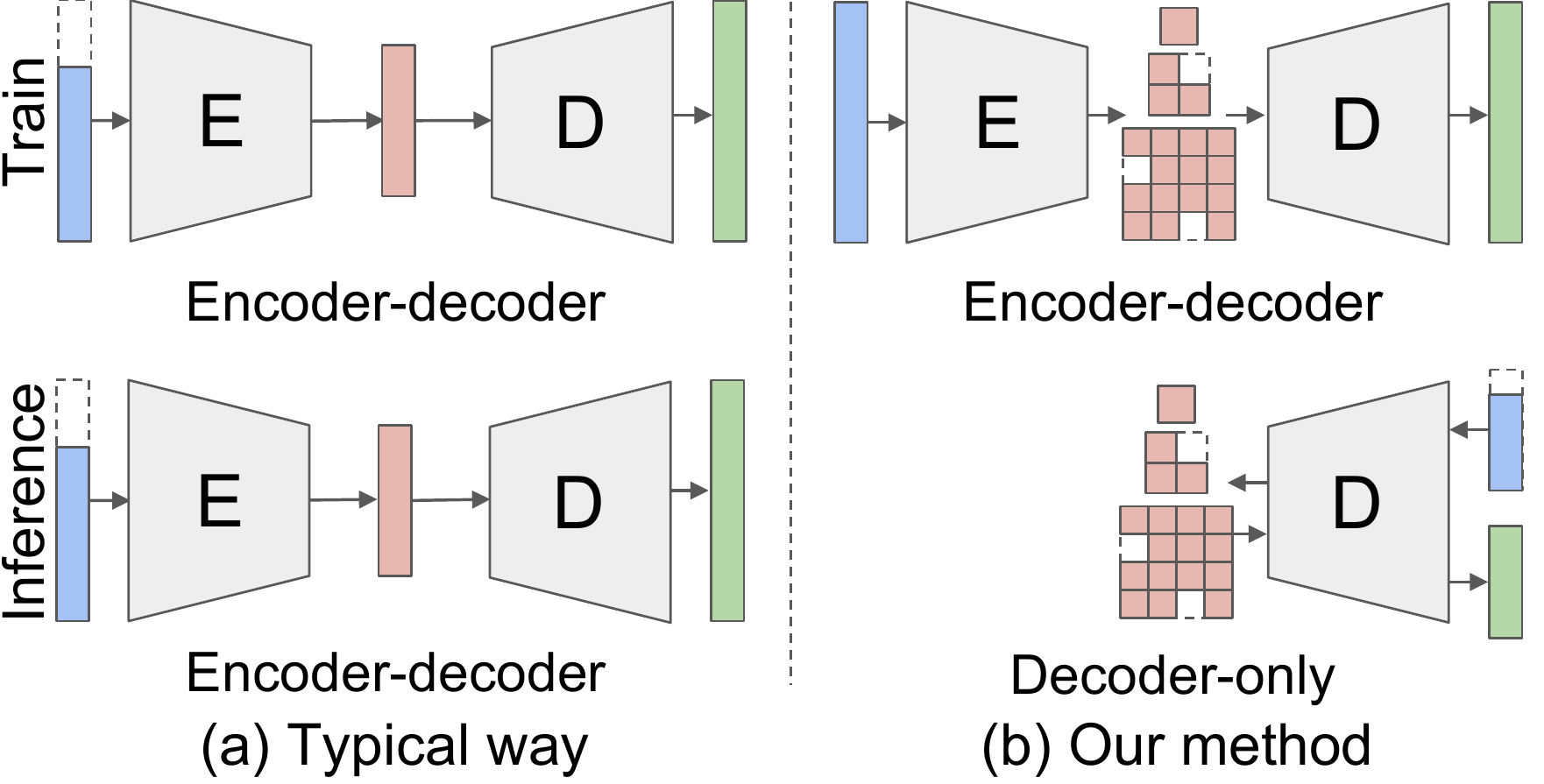}
\vspace{-0.5em}
\caption{(a) The conventional way of training an auto-encoder for completion is to feed partial data~(blue) from the encoder side. In this way, the encoder plays a crucial role in completion during inference. (b) We instead apply random dropout to our latent grids during training~(top), which forces the decoder to learn to complete the shape~(green). As a result, detailed completion can be achieved with decoder-only latent optimization~(bottom). For simplicity, we visualize levels of decoders as one block.}
\vspace{-1.5em}
\label{fig:dropout}
\end{figure}

\paragraph{Latent grid dropout}
There are mainly two standard ways to make a deep implicit function model work for completion tasks. The more conventional way, as illustrated in~\Fig{dropout}~(a), is training the model to take partial data as input and complete them. In this manner, the completion functionality is distributed among the encoder and decoder, therefore different encoders need to be trained for different completion tasks.
Another way is decoder-only latent optimization, where the encoder is not needed during test-time and the latent code is optimized based on partial data~\cite{park2019deepsdf}. This manner provides higher accuracy on observed regions and directly generalizes to different completion modalities (depth image, partial scan, etc.) without retraining. 
However, it only works for global methods and cannot be applied to local methods. The reason is that for unobserved regions with no data point, the corresponding local latent codes cannot be optimized and will stay as initialization. Such latent codes would then be decoded into wrong shapes by the decoder.

To address this, we propose to train with complete shapes, but apply random dropout to latent grids, as shown in~\Fig{dropout}~(b). The motivation is to simulate partial data in the latent space rather than the input space, hence forcing the decoder to learn to complete shapes without encoder. Specifically, for each level $\level>0$, we apply spatial dropout to $\latentgrid_n$, but keep the full content of $\hat\latentgrid_n$, so that the decoder can utilize the global context from level $0$. Note that our proposed multi-level architecture and \textit{global connection} make this dropout strategy possible during training: this cannot be applied to other global or local approaches, without substantial changes in the architectures.

\subsection{Inference}
\label{sec:inference}
We discuss our inference process with respect to auto-encoding (complete observation) and shape completion (partial observation).

\paragraph{Auto-encoding} \MDIF supports both encoder-decoder inference and decoder-only latent optimization. For applications that emphasize efficiency, encoder-decoder inference is a better choice, as it only has one feed-forward pass. For applications that require accuracy, decoder-only latent optimization is preferred.

\paragraph{Shape completion} Here we focus on shape completion from a single depth image via decoder-only latent optimization, due to its benefits in accuracy and generalizability.
We initialize all latent codes as zeros. Similar to global methods, level 0 can be optimized to have a coarse but complete reconstruction. For higher levels, the decoder is trained to add detail onto the observed parts, while produce sparse residual to the unobserved part.
For this optimization process to work, we need to properly sample points and modify the loss function to accommodate incomplete observation.

When sampling the point set $\pointset$ from a depth image, since part of the shape is occluded, we cannot simply sample points in the full volume as in training. Instead, we apply raycasting to sample camera-observable points as $\visiblesamples$, and occluded points as $\occludedsamples$. For level $\level=0$, the loss function is the same as~\Eq{loss_train} except only applied to visible points $\visiblesamples$. For level $\level>0$, the loss function $\loss_{\level}$ is modified to contain two terms as follows:
\begin{equation} 
\begin{split}
\loss_{\level} &= \loss_{\level}^{V} + \consistencyweight \loss_{\level}^{O},\\
\loss_{\level}^{V} &= \frac{1}{|\visiblesamples|}\sum_{\point \in  \visiblesamples}\left|\SDF_{\level}(\point)- \gt{\SDF}_{\level}(\point)\right|,\\ 
\loss_{\level}^{O} &=  \frac{1}{|\occludedsamples|}\sum_{\point \in \occludedsamples}(1-\Gaussian_{\sigma}\left(\distancefunction(\point, \visiblesamples))\right)\left|R_n(\point)\right|.
\end{split}
\label{eq:loss_lopt}
\end{equation}
The first term $\loss_{\level}^{V}$ is to minimize the difference between aggregated SDF prediction and ground truth for visible points. The second term $\loss_{\level}^{O}$ is for regularizing the residual of occluded points, such that the global shape from level 0 will be preserved for the unobserved part. In particular, $\distancefunction(\point, \visiblesamples)$ measures the closest distance from an occluded point $\point$ to the visible point set $\visiblesamples$, and is normalized by a Gaussian $\Gaussian$ of standard deviation $\sigma$. In practice, we empirically set $\consistencyweight=10$ and $\sigma=0.1$. We call the second term \textit{global consistency}.

%% file: 5_exp.tex
\section{Experiments}
\label{sec:eval}
In this section, we first validate the benefits of our proposed components by ablating important aspects (\Sec{ablation}). Then to evaluate the effectiveness of our approach, we compare with state-of-the-art methods on auto-encoding 3D shapes (\Sec{exp_ae}) and applications including point cloud completion (\Sec{point_cloud_completion}), voxel super-resolution (\Sec{voxel_super_resolution}) and shape completion from depth image (\Sec{shape_completion_depth}). These experiments demonstrate the capability of our method under different tasks and inference modes. We use 5 levels for \MDIF in the experiments and set the dimensions of the latent grids $\{\latentgrid_n\}$ as: $[1^3\times512, 2^3\times64, 4^3\times32, 8^3\times16, 16^3\times8]$. But note that \MDIF is flexible to use any number of levels.
During decoder-only latent optimization, we fix all other network parameters and only optimize over $\{\latentgrid_n\}$. Please refer to supplementary for more implementation details.

\subsection{Dataset \& Metrics}
Following prior works~\cite{jiang2020local, Genova_2020_CVPR}, we run the experiments on the ShapeNet dataset~\cite{shapenet2015} with train/test splits from 3D-R$^2$N$^2$~\cite{choy20163d}, which contain a subset of 13 categories in ShapeNet. We use all 13 categories in our experiments except for ablation studies where we only use the chair category. In all experiments, we only take the train split for training and leave out the test split for evaluation. For metrics, we use the \textit{Chamfer L2 distance} and \textit{F-Score} with the exact settings as in~\cite{Genova_2020_CVPR}. Since the Chamfer distance measures the average errors of all points, while the F-Score measures the ratio of good predictions, these two metrics do not always agree with each other: a better F-Score with a higher Chamfer distance usually indicates a few outliers resulting in significant error.

\subsection{Ablation Study}
\label{sec:ablation}
We conduct our ablation studies on the chair category of ShapeNet, for it contains large number of instances as well as significant intra-class shape variance. The models are all trained under encoder-decoder scheme and use decoder-only latent optimization during inference.

\begin{figure}[t]
\centering
\includegraphics[trim={0cm 3.25cm 9.5cm 0cm},clip,width=.8\linewidth]{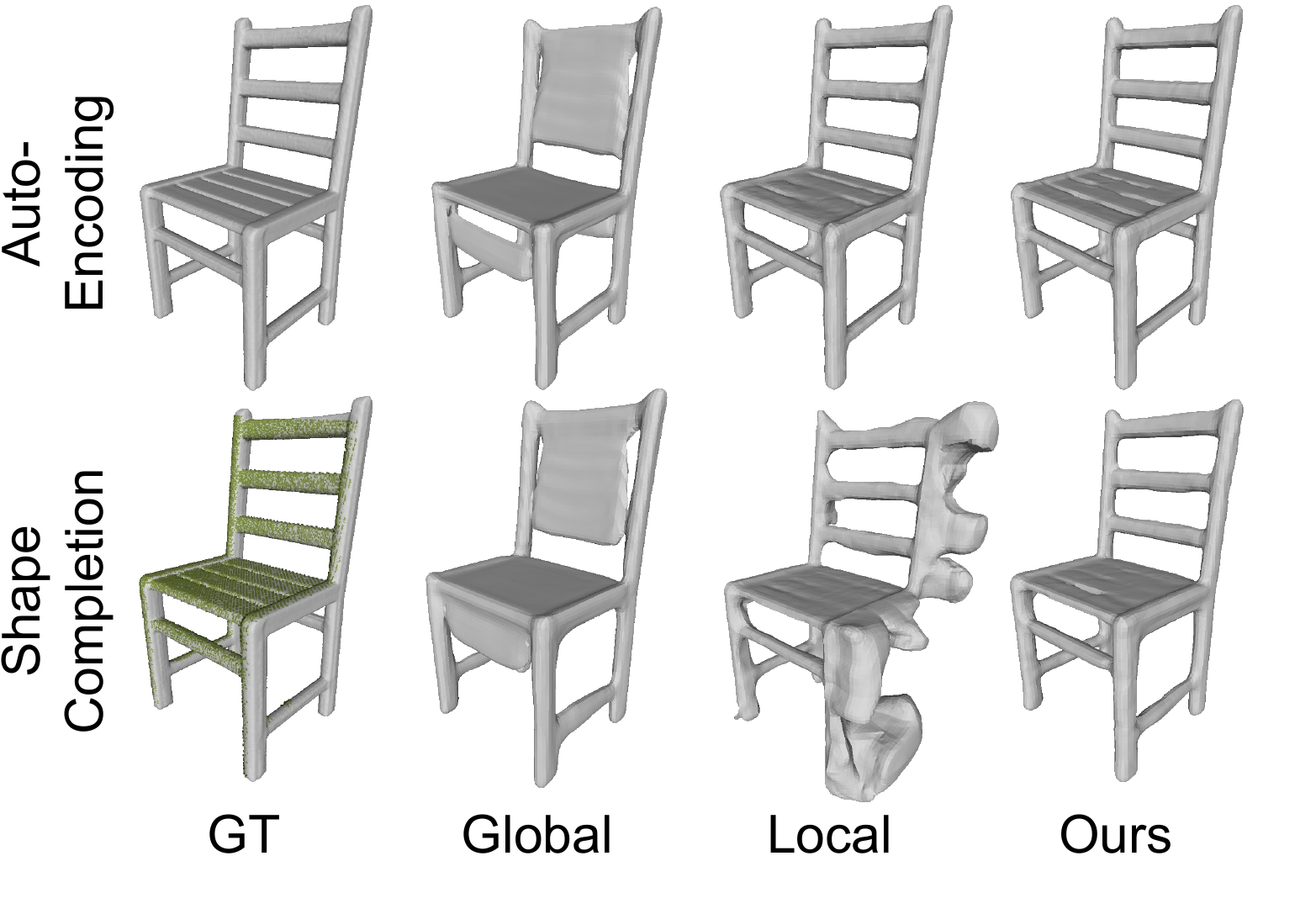}
\vspace{-1.25em}
\caption{\textbf{Qualitative comparisons among global/local/hierarchical baselines.} The global baseline lacks detail but behaves reasonably in both applications. The local method works well on observed data~(green dots) but generates noisy shapes for unobserved part. Our method has superior performance in both scenarios.}
\vspace{-1.5em}
\label{fig:architecture}
\end{figure}

\paragraph{Global/local/hierarchical} We compare \MDIF with a global and a local baselines to emphasize the impact of \MDIF's hierarchical model. The global baseline only has level 0, whilst the local baseline has only level 4 (a $16^3$ latent grid). In~\Table{architecture}, we compare with the baselines in terms of auto-encoding and shape completion from depth image. For auto-encoding, the local baseline clearly outperforms global, since it has larger capacity and the capability to capture details. On the flip side, for shape completion, the global baseline has better accuracy because the local baseline behaves randomly on the unobserved part, as visualized in the column 3 of ~\Fig{architecture}. Our \MDIF however, incorporates the benefits of global and local levels, and produces superior results in both tasks.

\begin{table}[t]
\centering
\resizebox{\linewidth}{!}{
\addtolength{\tabcolsep}{-2pt}  
\begin{tabular}{lcccccc}
\toprule
\multirow{2}{*}{Method} & \multicolumn{2}{c}{Auto-encoding} & \multicolumn{2}{c}{Shape Completion}\\
    & {Chamfer~($\downarrow$)} & {F-Score~($\uparrow$)} & {Chamfer~($\downarrow$)} & {F-Score~($\uparrow$)}\\
\midrule
Ours & \textbf{0.009} & \textbf{99.5} & \textbf{1.34} & \textbf{66.5} \\
Global baseline & 0.228 & 88.7 & 1.56 & 63.7 \\
Local baseline & 0.012 & 99.2 & 5.47 & 48.3 \\
\bottomrule
\end{tabular}
\addtolength{\tabcolsep}{2pt}
} %
\vspace{-0.75em}
\caption{\textbf{Quantitative comparisons among global/local/hierarchical baselines.} The local baseline has better auto-encoding performance than global, but performs poorly for shape completion from depth image. Our method combines the benefits of both.}
\vspace{-0.75em}
\label{tab:architecture}
\end{table}

\paragraph{Network components} In~\Table{ablation_study}, we incrementally compare the impact of four network components during decoder-only latent optimization. 

\begin{figure}[t]
\centering
\includegraphics[width=\linewidth]{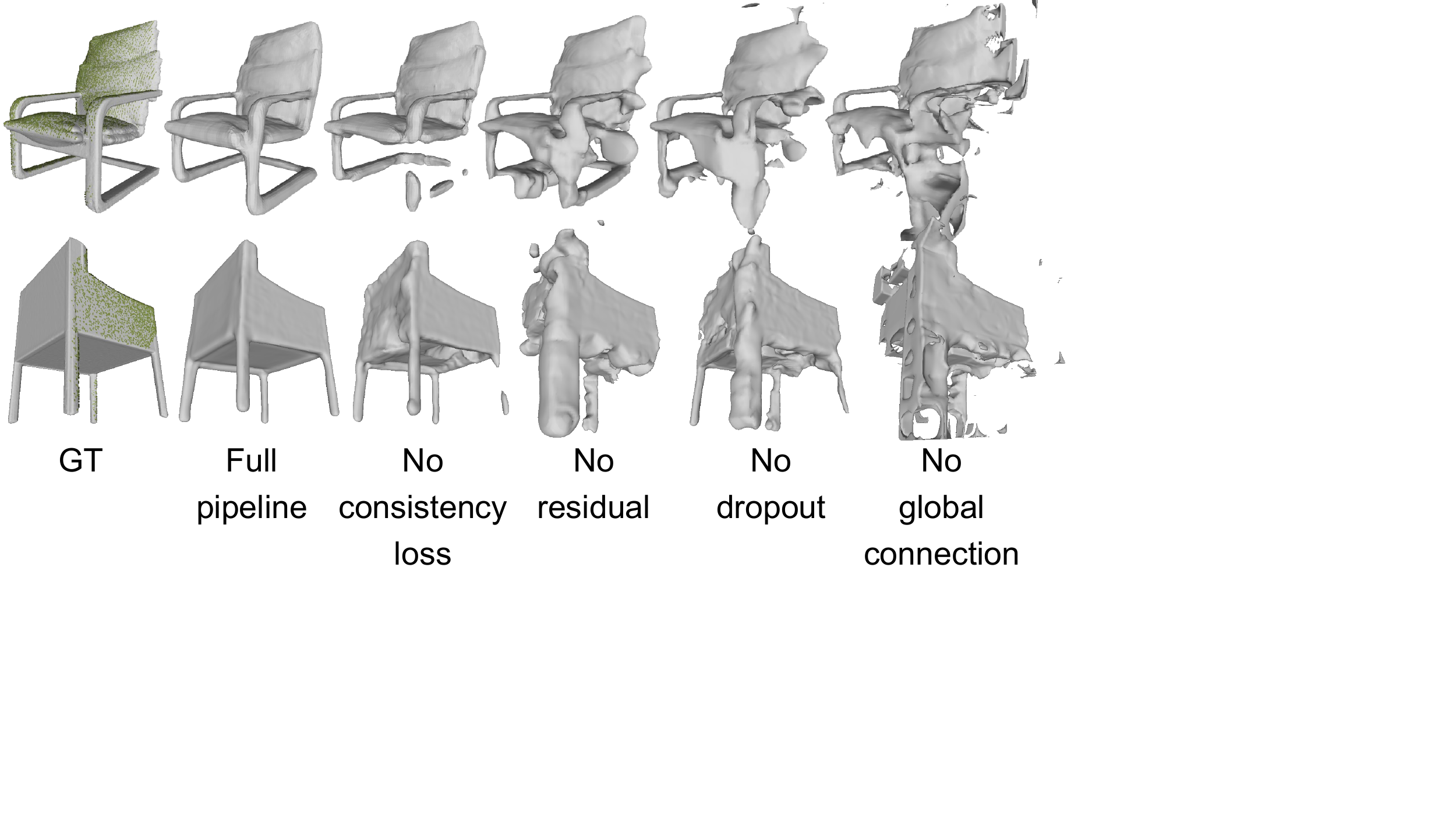}
\vspace{-1.5em}
\caption{\textbf{Qualitative ablation of the impacts of different components on shape completion from depth image.} Green dots are projected depth pixels~(observed data). Note that every component is necessary for good results.}
\vspace{-1.5em}
\label{fig:ablation_study}
\end{figure}

\textit{Global consistency} loss~(\Eq{loss_lopt}), which is designed to work for shape completion, has marginal improvements on the overall completion numbers. However, the column 3 of ~\Fig{ablation_study} shows that it is still important for clean reconstruction in unobserved regions.

We also compare the difference between decoding into SDF $\SDF$ or \textit{residual} $\residual$ in~\Eq{residual}. Since predicting residual forces lower levels to focus on the addition of fine detail, it is a stronger constraint and improves both auto-encoding and shape completion.

\textit{Latent grid dropout} is another component that is tailored to shape completion. Without it, the Chamfer error drastically increases from $3.0$ to $8.38$.
Also, it slightly improves decoder-only auto-encoding.
We hypothesize it is because dropout improves the generalization
of the decoders at levels 1-4 to test data and reduces the ambiguity between levels.

Finally, \textit{global connection} passes the global shape prior to other levels. Without it, the completion results are almost unconstrained on the unobserved part. It also helps auto-encoding, since without it, we are asking the network to add more detail without knowing what has been predicted by the previous levels, which is not sensible.

\begin{table}[t]
\centering
\resizebox{\linewidth}{!}{
\addtolength{\tabcolsep}{-2pt}  
\begin{tabular}{lcccccc}
\toprule
\multirow{2}{*}{Method} & \multicolumn{2}{c}{Auto-encoding} & \multicolumn{2}{c}{Shape Completion}\\
    & {Chamfer~($\downarrow$)} & {F-Score~($\uparrow$)} & {Chamfer~($\downarrow$)} & {F-Score~($\uparrow$)}\\
\midrule
Full pipeline & \textbf{0.009} & \textbf{99.5} & \textbf{1.34} & \textbf{66.5} \\
No consistency loss & - & - & 1.43 & 64.7 \\
No residual & 0.025 & 98.2 & 3.00 & 53.0 \\
No dropout & 0.026 & 97.9 & 8.38 & 43.0 \\
No global connection & 0.086 & 93.9 & 19.9 & 39.6\\
\bottomrule
\end{tabular}
\addtolength{\tabcolsep}{2pt}
} %
\vspace{-0.75em}
\caption{\textbf{Quantitatively ablate the impacts of different components on auto-encoding and shape completion from depth image.}}
\vspace{-1.0em}
\label{tab:ablation_study}
\end{table}

\subsection{Auto-Encoding 3D Shapes}
\label{sec:exp_ae}

\begin{figure}[t]
\centering
\includegraphics[width=\linewidth]{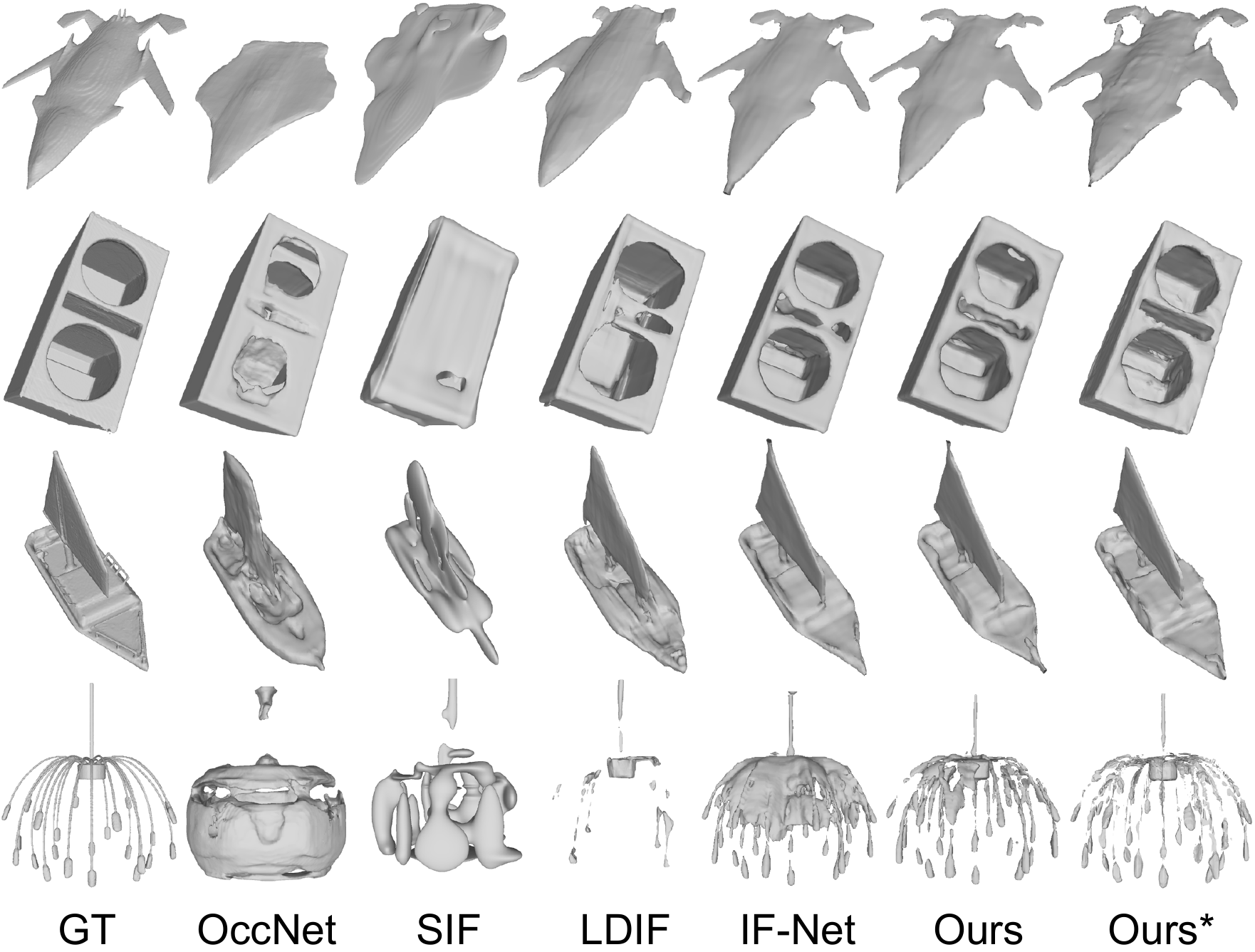}
\vspace{-1.75em}
\caption{\textbf{Auto-encoding results on test split.} Our method better reconstructs the groundtruth and recovers fine details. $*$: decoder-only latent optimization.}
\label{fig:qualitative_ae}
\vspace{-1.75em}
\end{figure}

\paragraph{Accuracy on test split} We first evaluate the auto-encoding accuracy under encoder-decoder inference for the test shapes in 3D-R$^2$N$^2$. We compare our approach with state-of-the-art DIF methods including OccNet (``Occ.'')~\cite{mescheder2019occupancy}, SIF~\cite{genova2019learning}, LDIF~\cite{Genova_2020_CVPR} and IF-Net (``IF.'')~\cite{chibane2020implicit}. The results for OccNet, SIF and LDIF are kindly provided by the authors of~\cite{Genova_2020_CVPR}. For IF-Net, it originally uses high-resolution latent grids (up to $128^3$) which altogether is over 20 times larger than the input grid ($128^3$) in the number of parameters. This would make the encoded latent grids meaningless for auto-encoding task. Therefore in this experiment, we constrain IF-Net to only use latent grids up to $16^3$ resolution (same as our approach) and have same total number of parameters in the latent grids as our approach. ~\Table{quantitative_ae} (middle columns) show the average metrics across 13 categories. Our method achieves slightly higher F-Score and much lower Chamfer error, which means it works better overall and on hard examples too. As visualized in~\Fig{qualitative_ae}, our method preserves details well and represents thin structures much better than the competing methods (see the last row).

Next, we evaluate the performance under decoder-only latent optimization. We compare with OccNet (``Occ.'')~\cite{mescheder2019occupancy}, IM-Net (``IM.'')~\cite{chen2019learning} and a local baseline (resembles~\cite{jiang2020local,chabra2020deep}), as shown in~\Table{quantitative_ae} (right columns). Our method also performs the best under this inference mode and can improve over encoder-decoder inference by a large margin. The last column of~\Fig{qualitative_ae} shows qualitative results.

\begin{figure}[t]
\centering
\includegraphics[width=1.0\linewidth]{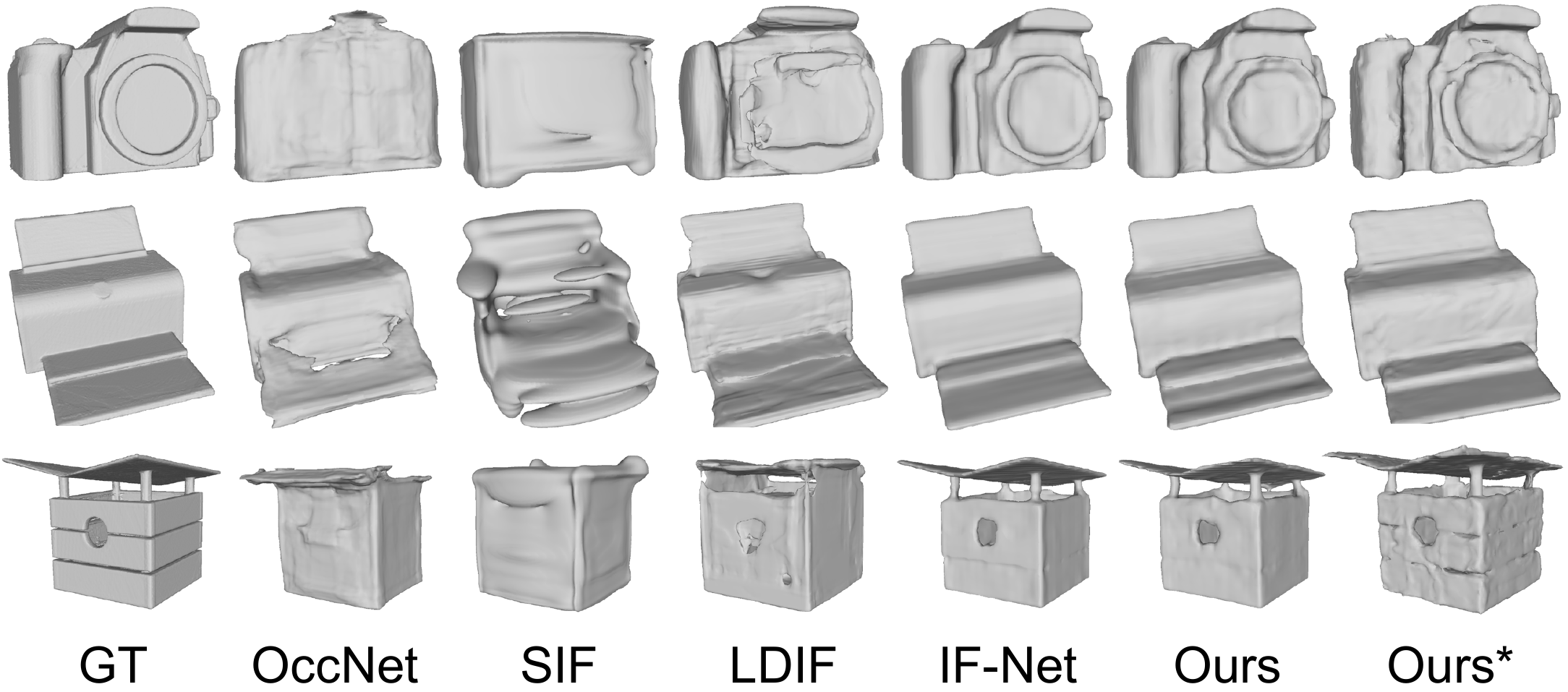}
\vspace{-1.75em}
\caption{\textbf{Auto-encoding results for unseen categories.} $*$: decoder-only latent optimization.}
\vspace{-0.5em}
\label{fig:unseen}
\end{figure}

\paragraph{Generalizability} In this experiment, we study the generalizability to shapes vastly different from training data. We test the trained models from the last experiment without fine-tuning on 10 ShapeNet categories that are unseen during training. In ~\Table{unseen}, we compare the performance under both inference modes, and our method respectively outperforms other methods. While global methods generalize poorly to unseen categories, our method performs equally well as seen categories. Qualitative results are shown in~\Fig{unseen}.

\begin{table}[t]
\centering
\resizebox{\linewidth}{!}{
\addtolength{\tabcolsep}{-3pt}  
\begin{tabular}{l|ccccc|cccc}
\toprule
 & Occ. & SIF & LDIF & IF. & Ours & Occ.* & IM.* & Local* & Ours* \\
\midrule
Chamfer & 0.49 & 1.18 & 0.4 & 0.39 & \textbf{0.19} & 0.43 & 0.46 & 0.14 & \textbf{0.10} \\
F-Score & 81.9 & 59 & 92.2 & 92.9 & \textbf{93.0} & 81.4 & 86.7 & 96.9 & \textbf{97.0} \\
\bottomrule
\end{tabular}
} %
\vspace{-0.75em}
\caption{\textbf{Auto-encoding accuracy for objects in 3D-R$^2$N$^2$ test set.} Middle columns compare methods under encoder-decoder inference while right columns compare under decoder-only latent optimization. $*$: decoder-only latent optimization.}
\vspace{-1em}
\label{tab:quantitative_ae}
\end{table}

\begin{figure}[t]
\begin{subfigure}{0.495\linewidth}
\begin{center}
\includegraphics[width=0.815\linewidth]{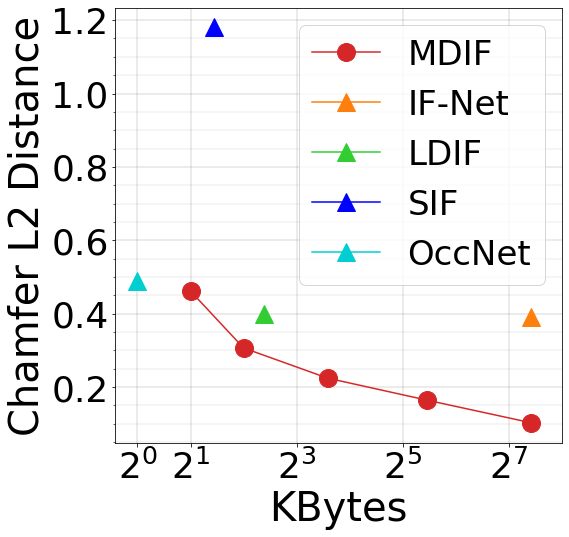}
\end{center}
\end{subfigure}
\begin{subfigure}{0.495\linewidth}
\begin{center}
    \includegraphics[width=0.815\linewidth]{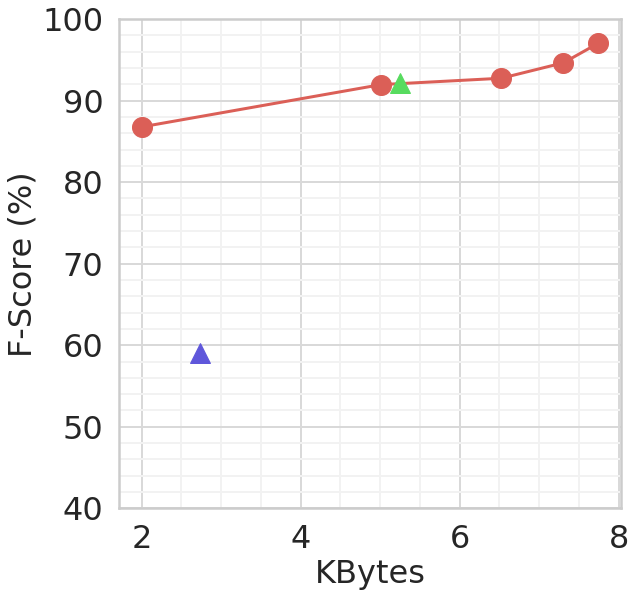}
\end{center}
\end{subfigure}
\vspace{-0.5em}
\caption{\textbf{Progressive refinement rate-distortion.} Our model allows progressive transmission of the latent codes of each level for refinement. This figure shows the accumulated latent code size (in bytes) and the respective distortion. For reference, the original $128^3$ volume is 8MB.}
\vspace{-1.5em}
\label{fig:progressive}
\end{figure}

\paragraph{Progressive refinement} One unique property of \MDIF is the capability to decode shapes in different levels of detail. This enables the progressive refinement application in graphics, where 3D data are encoded into different levels of detail and progressively rendered. Since \MDIF has a multi-level architecture, this can be easily achieved by only decoding the shape up until a certain level. 
\Fig{progressive} shows the distortion against the accumulated latent code size in bytes of each level, \ie, latent space capacity. \MDIF consistently improves with each level added. When under similar bytes, \MDIF still outperforms SIF, LDIF and IF-Net.

\begin{table}[t]
\centering
\resizebox{\linewidth}{!}{
\addtolength{\tabcolsep}{-3pt}
\begin{tabular}{l|ccccc|cccc}
\toprule
 & Occ. & SIF & LDIF & IF. & Ours & Occ.* & IM.* & Local* & Ours* \\
\midrule
Chamfer & 0.85 & 1.48 & 0.53 & 0.40 & \textbf{0.17} & 0.62 & 0.47 & 0.063 & \textbf{0.054} \\
F-Score & 66.6 & 43.0 & 84.4 & 92.4 & \textbf{92.8} & 71.1 & 80.5 & 97.5 & \textbf{97.5} \\
\bottomrule
\end{tabular}
} %
\vspace{-0.75em}
\caption{\textbf{Auto-encoding accuracy for objects in unseen categories.} Middle columns compare methods under encoder-decoder inference while right columns compare under decoder-only latent optimization. $*$: decoder-only latent optimization.}
\vspace{-0.75em}
\label{tab:unseen}
\end{table}

\subsection{Point Cloud Completion}
\label{sec:point_cloud_completion}
In this application, we take voxelized point cloud instead of SDF grid as input. We follow the same steps as IF-Net~\cite{chibane2020implicit} to produce such input: first sample 300 points from object surface and then voxelize these points into a $128^3$ grid. We compare our method with IF-Net, where both methods use encoder-decoder inference. As indicated in \Table{point_cloud_completion} (middle 2 columns), our method has higher F-Score and much lower Chamfer error. This reveals that our method is more accurate and stable in prediction. \Fig{pcc_vsr} (top row) shows results for one example data. Our method preserves the cavity in the legs while IF-Net incorrectly fills part of the cavity.

\subsection{Voxel Super-Resolution}
\label{sec:voxel_super_resolution}
\begin{figure}[t]
\centering
\includegraphics[width=0.825\linewidth]{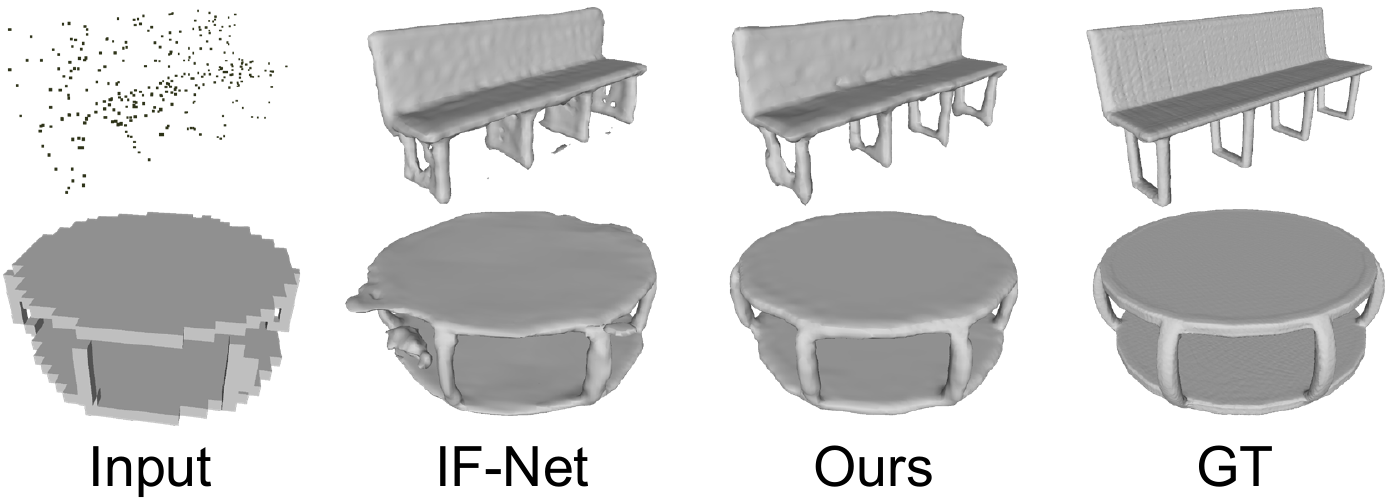}
\vspace{-0.75em}
\caption{Qualitative results for point cloud completion (top row) and voxel super-resolution (bottom row). More qualitative results are available in supplementary.}
\vspace{-1.5em}
\label{fig:pcc_vsr}
\end{figure}

In this task, we input $32^3$ occupancy grid and ask the network to predict the underlying continuous implicit field. The resolution of output grid for meshing is $128$. We compare our method with IF-Net, with both under encoder-decoder inference. \Table{point_cloud_completion} (right 2 columns) show the quantitative results. Similar to the case in point cloud completion, our method outperforms IF-Net with a large margin in Chamfer error. In \Fig{pcc_vsr} (bottom row), we show qualitative results on one example data. Our method is reasonably accurate in both global shape and local detail while IF-Net produces artifacts near the object boundary.

\begin{table}[t]
\centering
\resizebox{\linewidth}{!}{
\addtolength{\tabcolsep}{-2pt}  
\begin{tabular}{l|cc|cc}
\toprule
\multirow{2}{*}{Method} & \multicolumn{2}{c|}{Point Cloud Completion} & \multicolumn{2}{c}{Voxel Super-Resolution}\\
    & {Chamfer~($\downarrow$)} & {F-Score~($\uparrow$)} & {Chamfer~($\downarrow$)} & {F-Score~($\uparrow$)}\\
\midrule
IF-Net & 1.61 & 85.0 & 1.82 & 65.4 \\
Ours & \textbf{0.39} & \textbf{86.1} & \textbf{0.96} & \textbf{66.9} \\
\bottomrule
\end{tabular}
\addtolength{\tabcolsep}{2pt}
} %
\vspace{-0.75em}
\caption{Quantitative results for point cloud completion and voxel super-resolution.}
\vspace{-0.75em}
\label{tab:point_cloud_completion}
\end{table}

\subsection{Shape Completion from Depth Image}
\label{sec:shape_completion_depth}
\begin{figure}[t]
\begin{center}
\includegraphics[width=.575\linewidth]{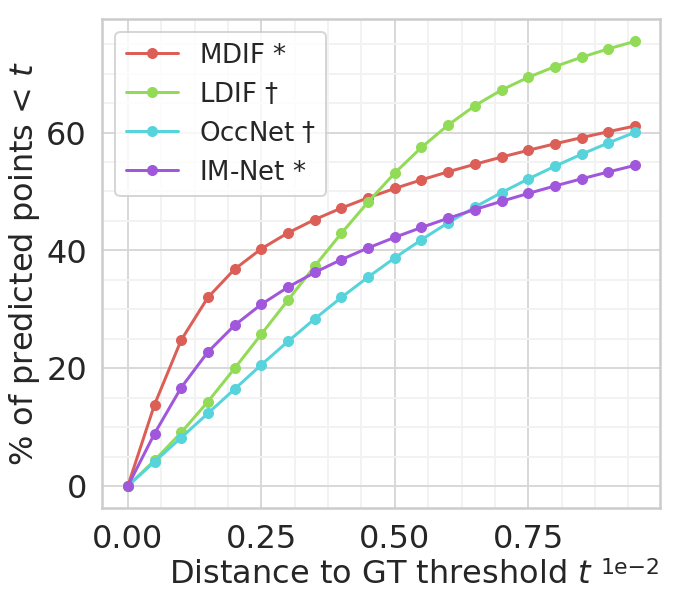}
\end{center}
\vspace{-1.5em}
\caption{\textbf{Shape completion from depth image.} The proportion of predicted points with distance to groundtruth smaller than different thresholds. $\dagger$: encoder-decoder inference; $*$: decoder-only latent optimization.}
\vspace{-10pt}
\label{fig:depth_comp_plot}
\end{figure}

\begin{figure}[t]
\centering
\includegraphics[trim={0cm 0cm 13.25cm 0cm},clip,width=0.85\linewidth]{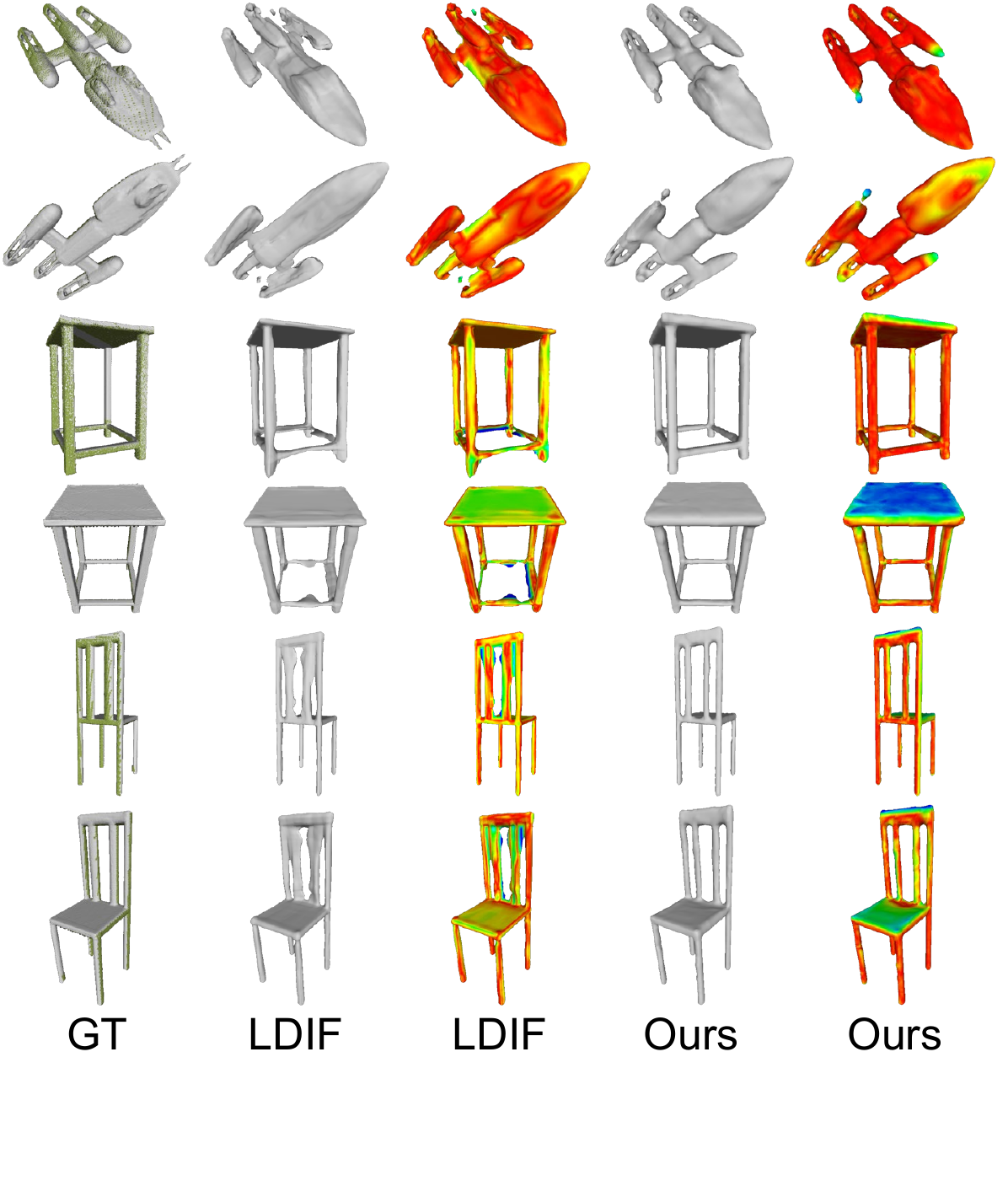}
\vspace{-3.25em}
\caption{\textbf{Qualitative results for shape completion from depth image.} We visualize the reconstruction and error maps (\textcolor{red}{low}/\textcolor{yellow}{mid}/\textcolor{blue}{high}) of three objects from two different angles.
In GT column, green dots represent observed parts.
}
\vspace{-1.5em}
\label{fig:depth_comp}
\end{figure}

Our final experiment investigates shape completion from depth image. We compare \MDIF with IM-Net, OccNet and LDIF. OccNet and LDIF use encoder-decoder inference while IM-Net and \MDIF use decoder-only latent optimization. Note that for IM-Net and \MDIF, we directly use the model trained in the auto-encoding task (\Sec{exp_ae}) without retraining or finetuning. This is considered a benefit of decoder-only latent optimization.
\Fig{depth_comp_plot} reports the percentages of surface points with distance to groundtruth smaller than different thresholds. \MDIF has a good proportion of points with low error and consistently outperforms IM-Net at all thresholds, reflecting its advantage on preserving details in observed regions. However, \MDIF has higher error in unobserved regions than methods under encoder-decoder inference (OccNet, LDIF). This is illustrated in~\Fig{depth_comp}, where the errors of our results are mostly on the occluded side. For example, in row 4 where the table top is completely unobserved, our estimation is thicker than groundtruth, hence resulting in higher error. Despite this, the predicted shape still looks plausible. 
This and other examples suggest that the Chamfer distance and F-Score are suited for assessing the observed parts, but not for the unobserved parts where many plausible solutions exist. 
Therefore, to evaluate plausibility, we further conduct a user study that votes between \MDIF and LDIF results on 32 pairs of examples (please refer to supplementary for details). The results show that 54.2\% of the participants chose MDIF results as more plausible, whilst 31.9\% thought LDIF results were better. In addition, 13.9\% could not decide between MDIF and LDIF. Moreover, when compared with the quantitative
metrics, 68.1\% disagree with the Chamfer distance, and 51.4\% disagree with the F-Score.

%% file: 6_conclusion.tex
\section{Conclusion}
In this paper, we present MDIF, a multi-resolution deep implicit function to progressively represent and reconstruct geometries. MDIF is trained end-to-end in an encoder-decoder fashion and supports both encoder-decoder inference and decoder-only latent optimization.
We demonstrate that MDIF outperforms state-of-the-art methods on tasks including auto-encoding 3D shapes, point cloud completion and voxel super-resolution. We further show that MDIF enables detailed decoder-only shape completion from a depth image: the details in observed regions are accurately preserved while the unobserved regions are completed with plausible shapes.
In the future, we would like to explore transferring details from observable parts to occluded parts in completion tasks. We also plan to apply \MDIF to more applications such as shape manipulation.

%% file: 7_supp.tex
\begin{figure*}[b]
\centering
\includegraphics[trim={0cm 0cm 0cm 0cm},clip,width=1.0\textwidth]{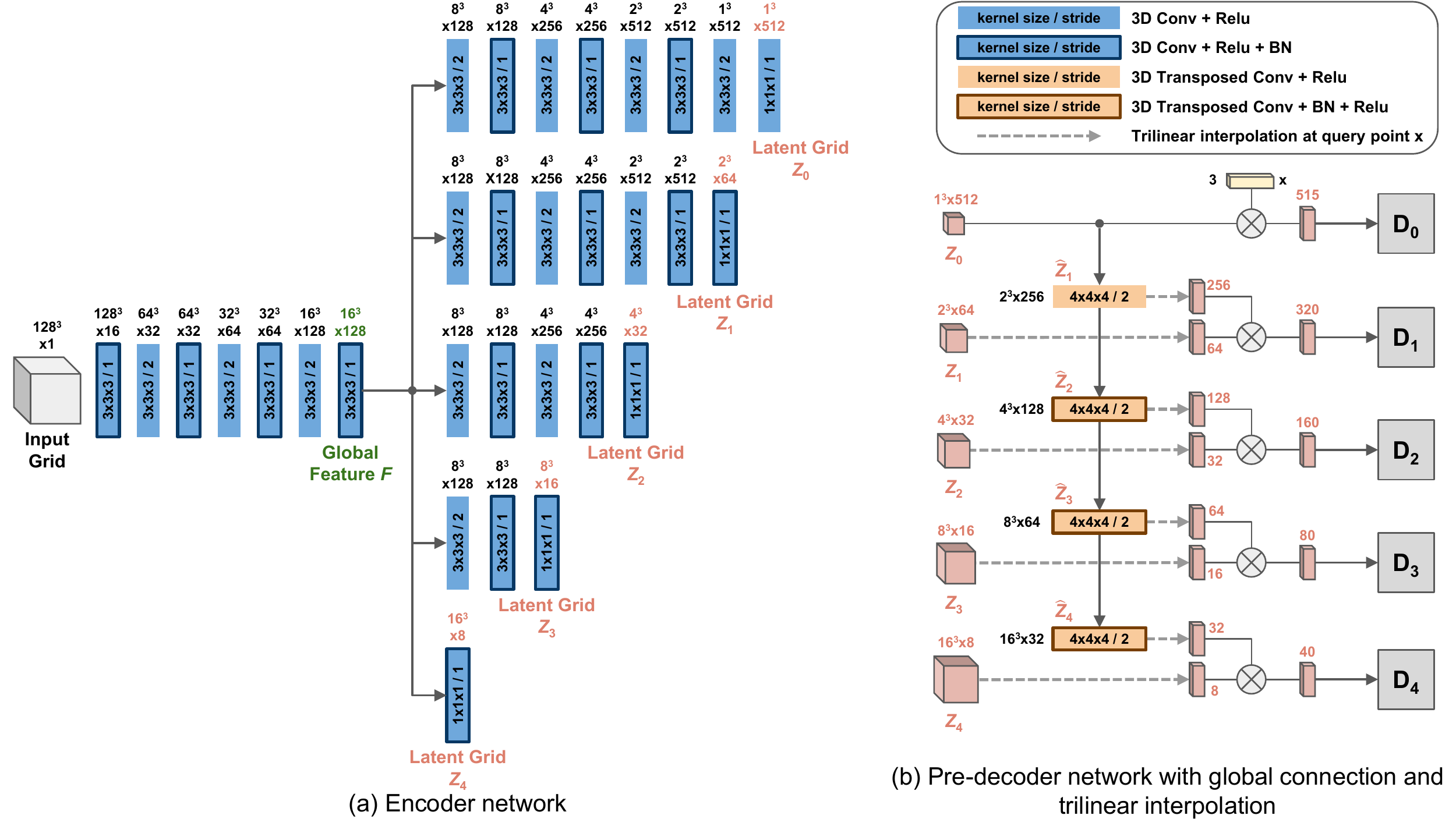}
\vspace{-1em}
\caption{Detailed architecture of our network.}
\label{fig:network_details}
\end{figure*}

\section{Supplementary Material}

\subsection{Implementation Details}
\paragraph{Detailed network architecture} \Fig{network_details} shows the detailed architecture of our network. On the left, \Fig{network_details}~(a) is the encoder network that is used in training and encoder-decoder inference. It takes 3D grid as input and outputs the latent grid $\latentgrid_n$ of each level. For the voxel super-resolution experiment (\Sec{voxel_super_resolution}), since the input is only $32^3$, we accordingly remove the first 4 convolution layers along with their activation and normalization layers.

On the right, \Fig{network_details}~(b) is the pre-decoder network. With latent grids $\{\latentgrid_n\}$ as input, it includes global connection and trilinear interpolation. The global connection consists of 3D transposed convolution layers to propagate global context from level 0 to other levels. Trilinear interpolation is utilized to obtain the latent codes at each query point, which are then fed into the decoders at each level. For level 0, the 3D position of query point is also fed into the decoder.
For the decoder modules, we use the same IM-Net~\cite{chen2019learning} architecture for each level, with the only difference in the input dimension. 

\clearpage
\paragraph{Hyperparameters} We implement our method in TensorFlow. During training, we set batch size as $8$ and train our network end-to-end. We use Adam as optimizer, with $\beta_1=0.9$, $\beta_2=0.999$ and a learning rate of $1\mathrm{e}{-4}$. 
The latent grid dropout rate is set as $0.5$ for the models that need to carry out decoder-only latent optimization while it is set as $0$ for the models that only run encoder-decoder inference (\eg, the models for point cloud completion and voxel super-resolution).

During decoder-only latent optimization, we optimize over $\latentgrid_n, n=0,1,...,4$ and keep other parameters fixed. We use Adam with the same configuration of $\beta_1, \beta_2$ as training, but at a higher learning rate of $1\mathrm{e}{-2}$ to accelerate convergence. In all our experiments, we only run latent optimization for $1000$ steps. For each step during auto-encoding, we randomly draw $2048$ points. For each step during shape completion, we randomly draw $2048$ camera-observable points, along with $1024$ occluded points for the \textit{global consistency loss}. 

\paragraph{Experiment details} For the training data, we use the watertight ShapeNet meshes from OccNet~\cite{mescheder2019occupancy} and normalize into bounding box with side length $1.28$. We also truncate SDF values at $0.05$.

For the auto-encoding experiment (\Sec{exp_ae}), as mentioned in the paper, IF-Net~\cite{chibane2020implicit} originally uses high-resolution latent grids which contain more parameters than the input grid. We therefore constrain IF-Net to only use latent grids with dimensions: $[8^3\times22, 16^3\times8]$. The resulting total number of parameters in the latent grids is the same as \MDIF.

For the point cloud completion (\Sec{point_cloud_completion}) and voxel super-resolution (\Sec{voxel_super_resolution}) experiments, unlike auto-encoding, the goal is to infer missing data rather than learn a compact latent space. Therefore, in these experiments, we use the original implementation of IF-Net which exploits high-resolution latent grids. Similarly, for \MDIF in these experiments, we additionally interpolate features at query points from high-resolution feature grids and feed into the decoders.

\subsection{Encoder-Decoder vs. Decoder-Only Inference}
\begin{figure}[h]
\centering
\includegraphics[trim={0cm 3.3cm 16cm 0cm},clip,width=0.5\textwidth]{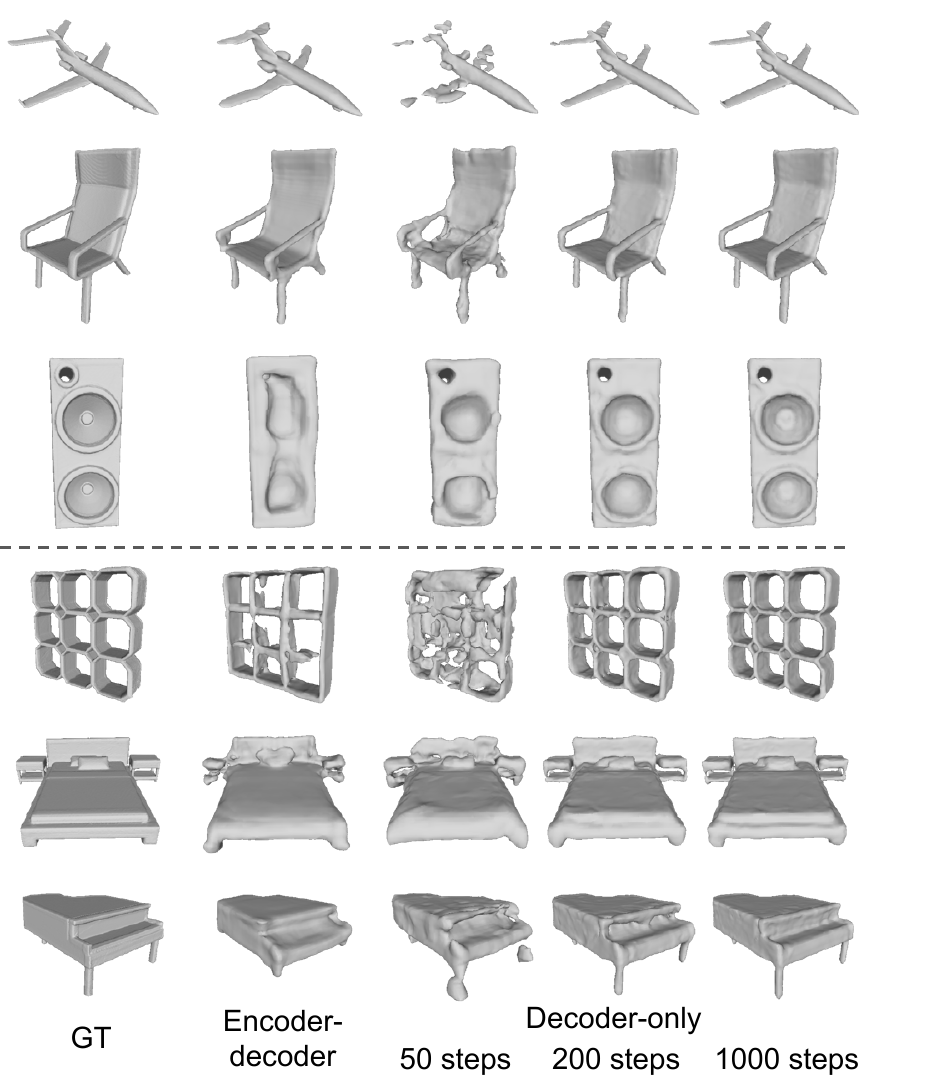}
\vspace{-1em}
\caption{\textbf{Encoder-decoder vs. decoder-only inference.} Auto-encoding results of \MDIF under encoder-decoder inference and decoder-only latent optimization. Top 3 rows: objects in 3D-R$^2$N$^2$ test split. Bottom 3 rows: objects in unseen categories.}
\label{fig:auto_enc-ff_vs_lopt_process}
\end{figure}

In \Fig{auto_enc-ff_vs_lopt_process}, we show qualitative auto-encoding results of \MDIF using encoder-decoder inference and decoder-only latent optimization. Compared with encoder-decoder inference, decoder-only latent optimization already produces more accurate reconstruction with only $200$ optimization steps. More steps further lower the error.

\subsection{Illustration of Ablation Baselines}
\begin{figure}[h]
\centering
\includegraphics[trim={0cm 0.5cm 6.5cm 0cm},clip,width=1.0\linewidth]{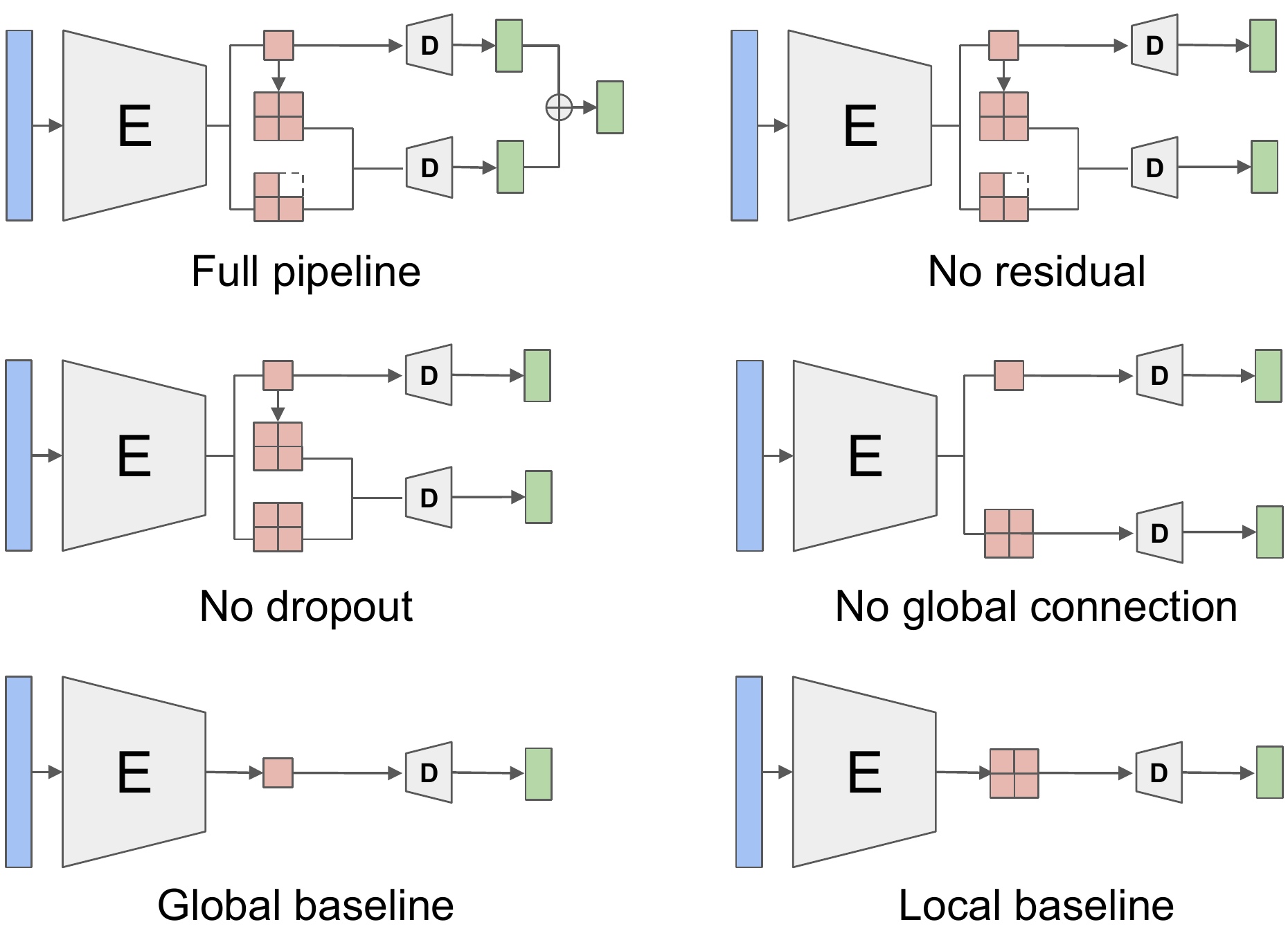}
\vspace{-1em}
\caption{\textbf{Illustration of ablation study baselines.} E: encoder; D: decoder.}
\vspace{-1em}
\label{fig:ablation_baselines}
\end{figure}

In \Fig{ablation_baselines}, we illustrate the baselines that we ablate in \Table{architecture} and \Table{ablation_study}.

\subsection{Comparison of Dropout and Consistency Loss}
\begin{figure}[t]
\centering
\includegraphics[trim={0cm 6cm 11cm 0cm},clip,width=0.5\textwidth]{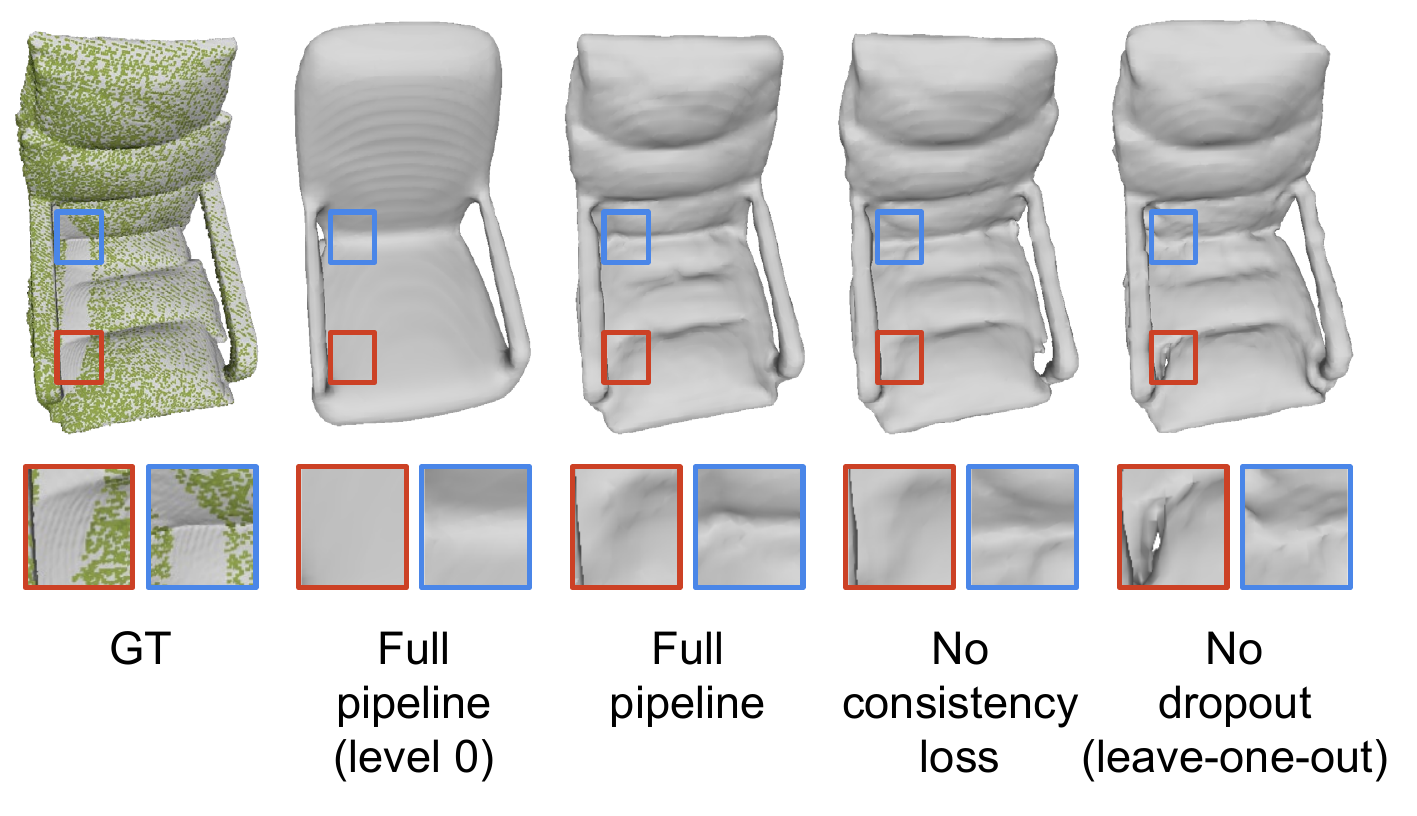}
\vspace{-2em}
\caption{\textbf{Ablation on latent grid dropout and consistency loss for the task of shape completion.} Green dots are observed depth points. Compared to the global consistency loss which regularizes regions far from observed points, latent grid dropout reduces noisy residuals and enables plausible detail synthesis on regions that are close to the observed part.}
\label{fig:ablation_dropout_vs_consistency}
\end{figure}

To further analyze the different contribution of \textit{latent grid dropout} and \textit{global consistency loss} on shape completion, we carry out a leave-one-out ablation on dropout where the only difference with full pipeline is the removal of latent grid dropout. Same as the baselines in \Table{ablation_study}, this ablation is conducted on the chair category of ShapeNet. In \Table{ablation_study1}, we show that the removal of dropout leads to slightly larger decrease in quantitative performance than the removal of consistency loss. Meanwhile, dropout impacts qualitative results in a different way than the consistency loss. As shown in \Fig{ablation_dropout_vs_consistency}, when dropout is applied (the third and fourth columns from the left), the model is able to synthesize plausible details on the unobserved regions that are close to the observed part (see insets at the bottom). On the contrary, without dropout (the rightmost column), the model tends to produce noisy residuals (red inset) or add no detail due to the consistency loss (blue inset).

\begin{table}[t]
\centering
\resizebox{\linewidth}{!}{
\addtolength{\tabcolsep}{-2pt}  
\begin{tabular}{lccccc}
\toprule
\multirow{3}{*}{Method} & \multicolumn{2}{c}{Shape Completion}\\
    & {Chamfer~($\downarrow$)} & {F-Score~($\uparrow$)}\\
\midrule
Full pipeline & \textbf{1.34} & \textbf{66.5} \\
No consistency loss & 1.43 & 64.7 \\
No dropout (leave-one-out) & 1.43 & 63.9 \\
\bottomrule
\end{tabular}
\addtolength{\tabcolsep}{2pt}
} %
\caption{Quantitatively ablate the impacts of consistency loss and latent grid dropout on shape completion from depth image. }
\label{tab:ablation_study1}
\end{table}

\subsection{Failure Cases}
\begin{figure}[t]
\centering
\includegraphics[trim={0cm 1cm 10cm 0cm},clip,width=0.5\textwidth]{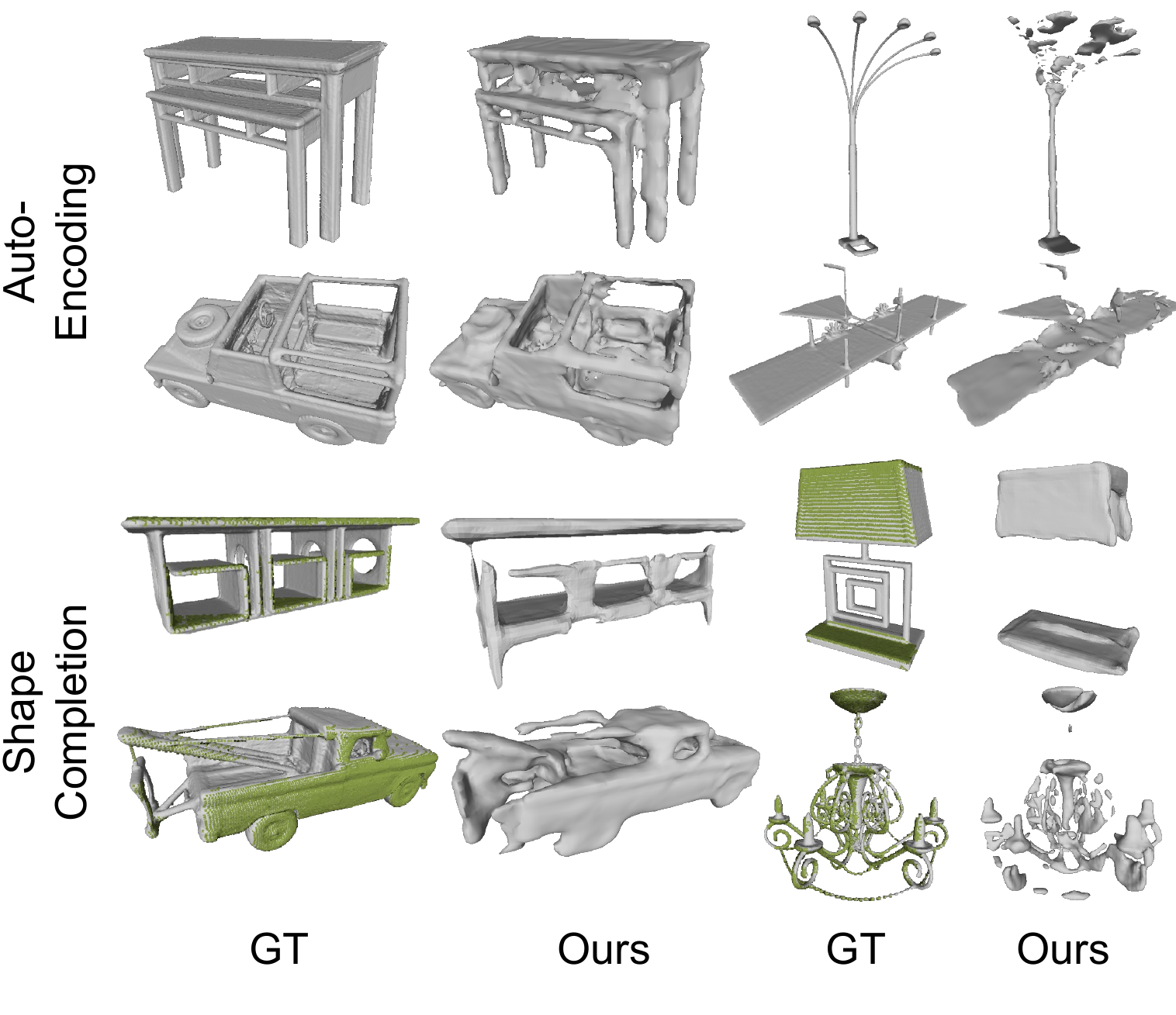}
\vspace{-2.0em}
\caption{\textbf{Failure cases.} Row 1 and 2: auto-encoding; Row 3 and 4: shape completion from depth image.}
\label{fig:failure_cases}
\end{figure}

\Fig{failure_cases} shows our failure cases under decoder-only latent optimization for auto-encoding and shape completion from depth image. For objects with very complex geometry or thin structures, our approach still faces challenges. For auto-encoding, such problems could be alleviated by using more levels and higher resolution latent grids. For shape completion, when an unobserved part~(\eg, the lamp body in row 3, column 3) is completely missing in the coarse prediction from level 0, our approach is unable to synthesize such delicate structures.

\subsection{Additional Ablation for Number of Levels}
In the paper, we use 5 levels as it is a good balance between accuracy and efficiency. But as previously indicated, MDIF is flexible to use other number of levels. 
In \Fig{progressive}, we showed progressive refinement rate-distortion for levels 1-5. 
Here in \Table{perf_vs_levels_ae}, we further show the auto-encoding accuracy under encoder-decoder inference with up to 8 levels.

\begin{table}[t]
\centering
\resizebox{\linewidth}{!}{
\addtolength{\tabcolsep}{-3pt}  
\begin{tabular}{l|cccc|cccc}
\toprule
 & Ours & Ours-6 & Ours-7 & Ours-8 & Ours & Ours-6 & Ours-7 & Ours-8 \\
\midrule
Chamfer & 0.19 & 0.13 & 0.13 & 0.12 & 0.17 & 0.14 & 0.13 & 0.13 \\
F-Score & 93.0 & 96.5 & 96.7 & 97.5 & 92.8 & 96.3 & 97.1 & 97.3 \\
\bottomrule
\end{tabular}
} %
\caption{\textbf{Auto-encoding accuracy with more levels.} Middle columns: 3D-R$^2$N$^2$ test set. Right columns: unseen categories. ``Ours" stands for 5 levels and ``Ours-$N$" stands for $N$ levels.}
\vspace{-0.5em}
\label{tab:perf_vs_levels_ae}
\end{table}

\subsection{Interpolation and Retrieval in Latent Space}
\begin{figure}[t]
\centering
\includegraphics[width=1.0\linewidth]{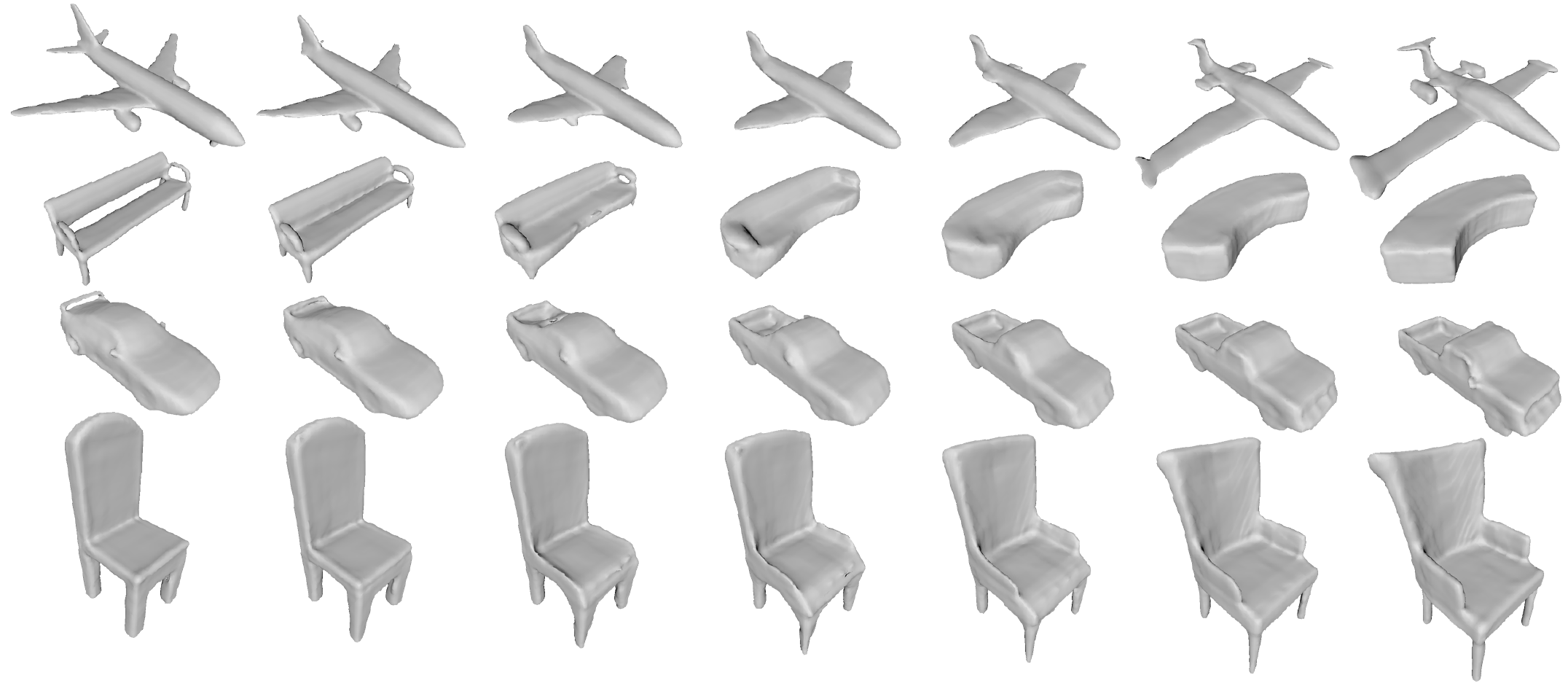}
\caption{\textbf{Linear interpolation in latent space.}}
\label{fig:latent_interp}
\end{figure}

\begin{figure}[t]
\centering
\includegraphics[width=1.0\linewidth]{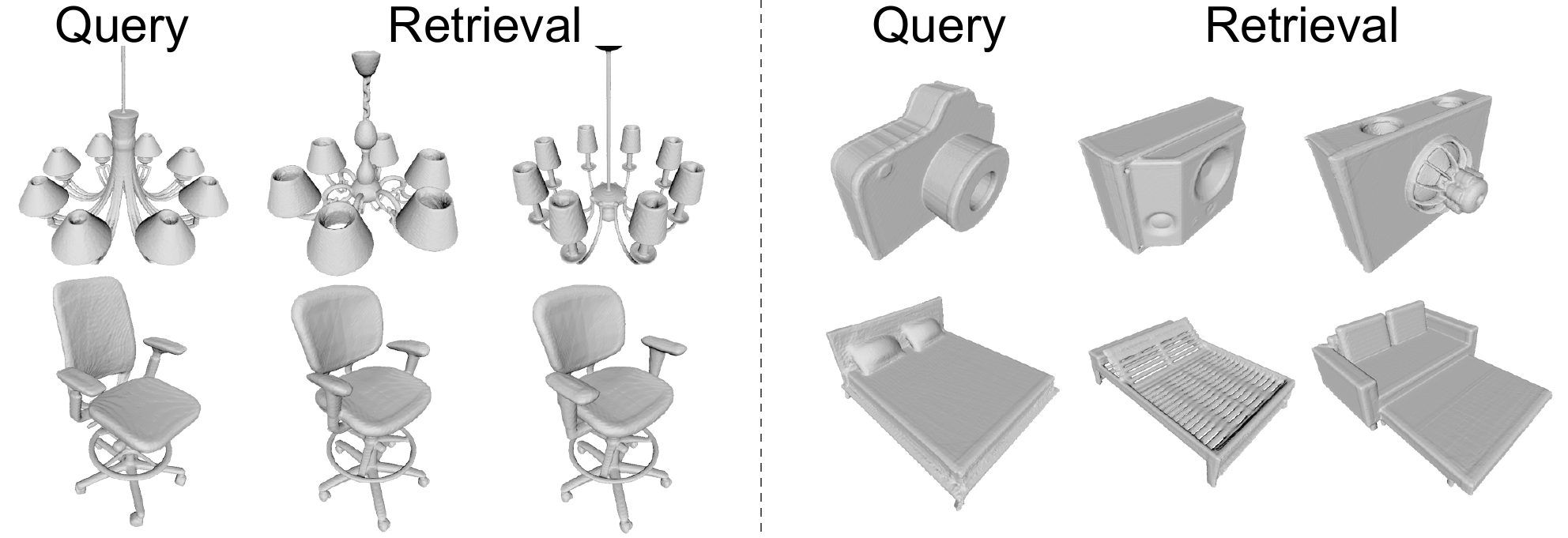}
\caption{\textbf{Object retrieval.} Queries are from test set (left) and unseen categories (right). Retrieved objects are from training set.}
\label{fig:object_retrieval}
\end{figure}

\Fig{latent_interp} shows linear interpolation in latent space. The latent codes for the two ends are obtained with encoder-decoder auto-encoding.
\Fig{object_retrieval} shows results for object retrieval based on latent codes (top-2 retrievals for each query object).

\subsection{Additional Qualitative Results}
\begin{figure}[t]
\centering
\includegraphics[trim={0cm 0cm 11.5cm 0cm},clip,width=1.0\linewidth]{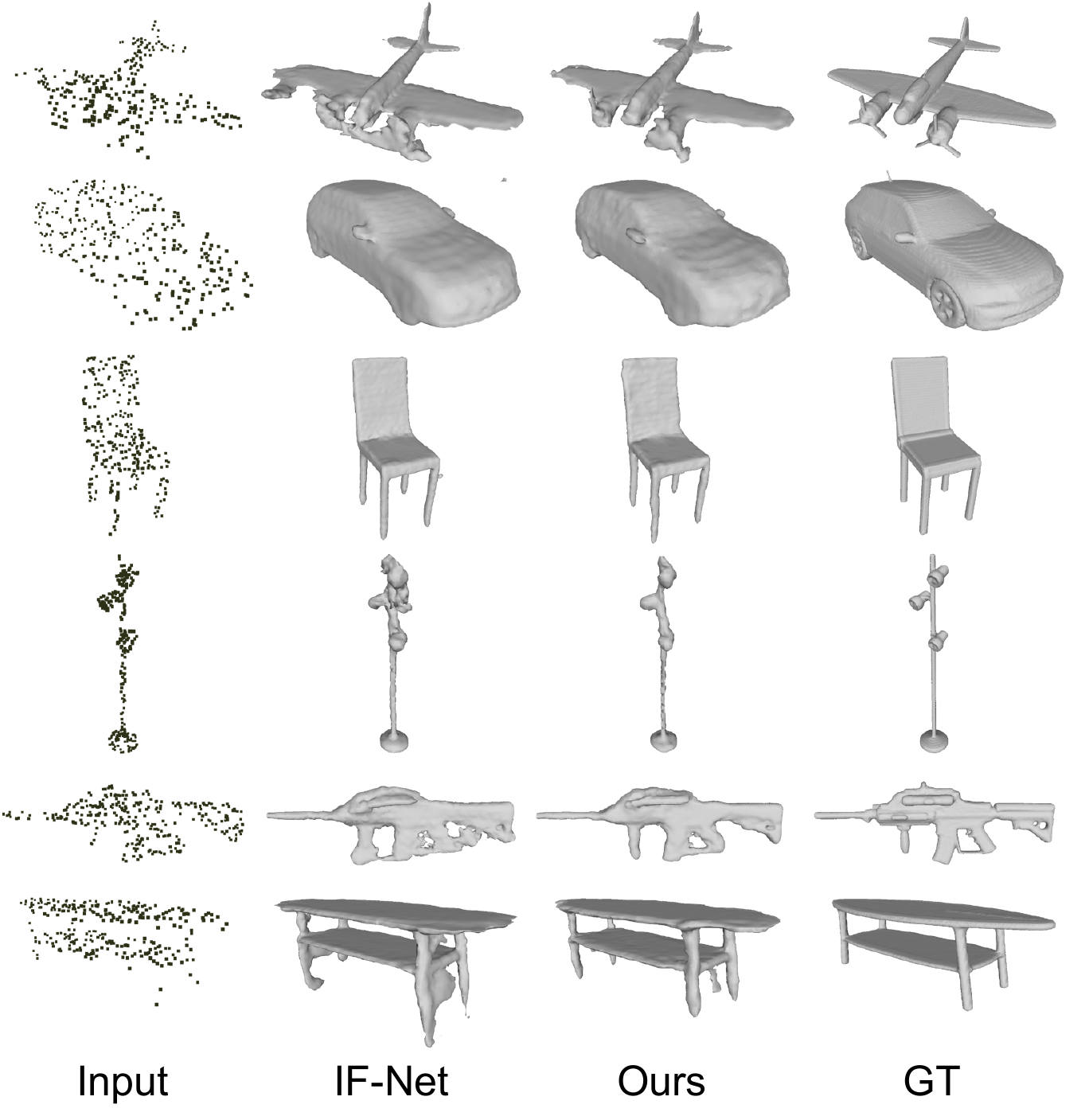}
\vspace{-1em}
\caption{\textbf{Point cloud completion}. Additional qualitative results.}
\label{fig:exp_pcc_supp}
\end{figure}

\begin{figure}[h]
\centering
\includegraphics[trim={0cm 0cm 11.5cm 0cm},clip,width=1.0\linewidth]{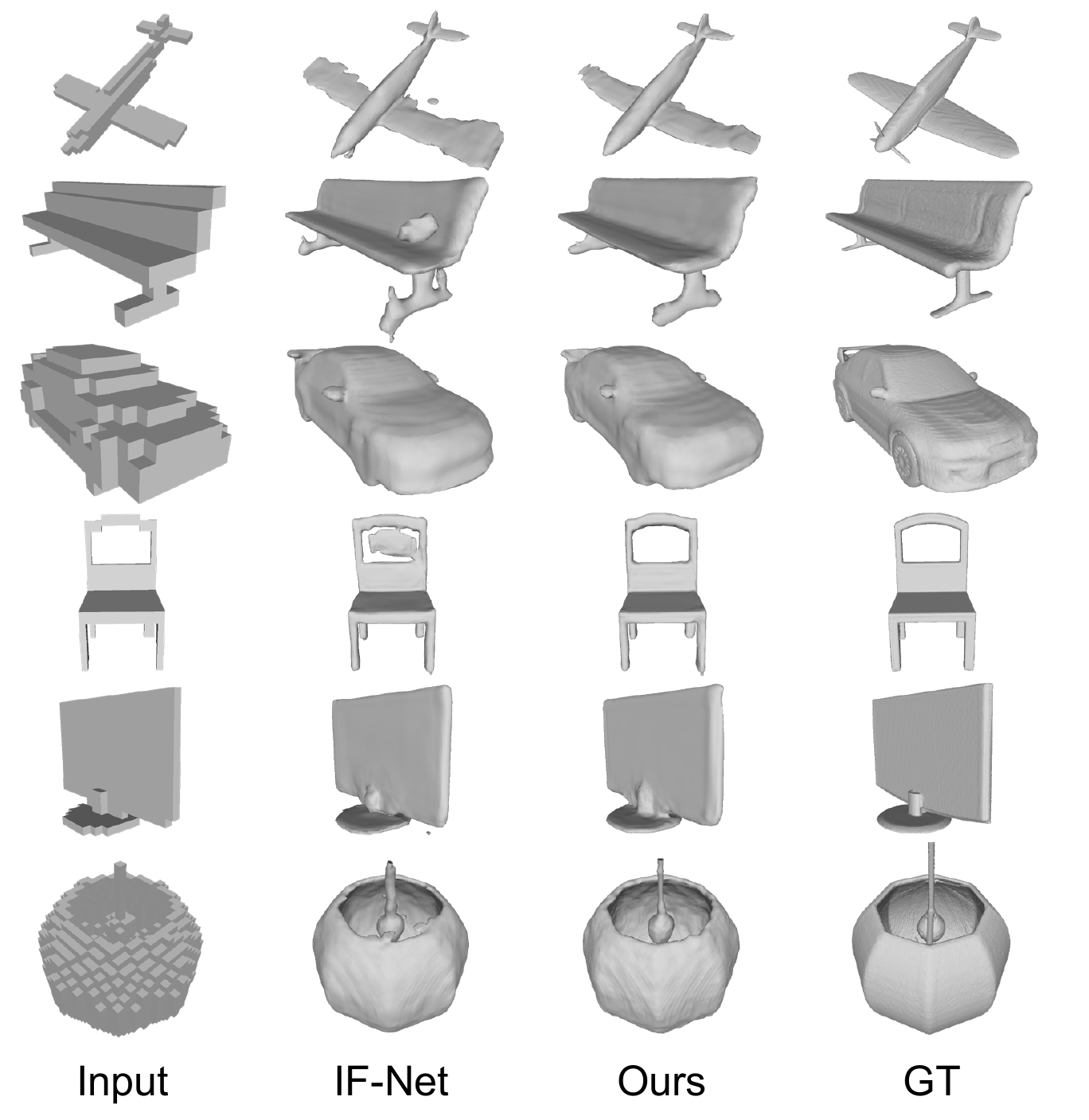}
\vspace{-1em}
\caption{\textbf{Voxel super-resolution}. Additional qualitative results.}
\label{fig:exp_vsr_supp}
\end{figure}

\Fig{exp_pcc_supp} and \Fig{exp_vsr_supp} show additional qualitative comparisons on point cloud completion and voxel super-resolution. Compared to IF-Net, our method generally produces cleaner reconstructions with less artifacts.

\subsection{Detailed Quantitative Results}
\Table{quantitative_ae_full} and \Table{unseen_ae_full} show per-category quantitative results (Chamfer L2 distance and F-Score) on auto-encoding. For encoder-decoder inference, we compare \MDIF with OccNet (``Occ.'')~\cite{mescheder2019occupancy}, SIF~\cite{genova2019learning}, LDIF~\cite{Genova_2020_CVPR} and IF-Net (``IF.'')~\cite{chibane2020implicit}. For decoder-only latent optimization, we compare \MDIF with OccNet (``Occ.'')~\cite{mescheder2019occupancy}, IM-Net (``IM.'')~\cite{chen2019learning} and a local baseline (resembles~\cite{jiang2020local,chabra2020deep}). 
\Table{pcc_vsr_full} shows per-category quantitative results (Chamfer L2 distance / F-Score) on point cloud completion and voxel super-resolution, where we compare \MDIF with IF-Net~\cite{chibane2020implicit} under encoder-decoder inference.

In these experiments, \MDIF has lower Chamfer errors for most categories and higher overall F-Score.

\begin{table*}[t]
\centering
\resizebox{\linewidth}{!}{
\addtolength{\tabcolsep}{-3pt}  
\begin{tabular}{l||ccccc|cccc||ccccc|cccc}
\toprule
\multirow{2}{*}{Category} & \multicolumn{9}{c}{Chamfer~($\downarrow$)} & \multicolumn{9}{c}{F-Score~($\uparrow$, \%)}\\
    & Occ. & SIF & LDIF & IF. & Ours & Occ.* & IM.* & Local* & Ours* & Occ. & SIF & LDIF & IF. & Ours & Occ.* & IM.* & Local* & Ours* \\
\midrule
airplane & 0.16 & 0.44 & 0.10 & 0.52 & \textbf{0.05} & 0.25 & 0.13 & 0.044 & \textbf{0.028} & 87.8 & 71.4 & 96.9 & 94.4 & \textbf{97.2} & 89.8 & 91.7 & 98.5 & \textbf{98.6} \\
bench & 0.24 & 0.82 & 0.17 & 0.31 & \textbf{0.08} & 0.34 & 0.22 & 0.121 & \textbf{0.052} & 87.5 & 58.4 & \textbf{94.8} & 92.6 & 92.4 & 85.2 & 88.6 & \textbf{96.0} & \textbf{96.0} \\
cabinet & 0.41 & 1.10 & 0.33 & \textbf{0.11} & 0.29 & 0.32 & 0.23 & 0.063 & \textbf{0.051} & 86.0 & 59.3 & 92.0 & \textbf{93.0} & 91.5 & 83.2 & 89.2 & \textbf{96.6} & \textbf{96.6} \\
car & 0.61 & 1.08 & \textbf{0.28} & 0.30 & 0.29 & 0.58 & 0.26 & 0.090 & \textbf{0.088} & 77.5 & 56.6 & 87.2 & \textbf{87.4} & 86.6 & 69.3 & 82.7 & \textbf{93.1} & 93.0 \\
chair & 0.44 & 1.54 & 0.34 & \textbf{0.10} & \textbf{0.10} & 0.38 & 0.43 & 0.042 & \textbf{0.035} & 77.2 & 42.4 & 90.9 & \textbf{94.5} & 93.8 & 80.2 & 82.5 & \textbf{97.7} & 97.6 \\
display & 0.34 & 0.97 & 0.28 & \textbf{0.07} & 0.08 & 0.35 & 0.20 & 0.043 & \textbf{0.019} & 82.1 & 56.3 & 94.8 & \textbf{96.1} & 95.1 & 82.3 & 89.4 & 98.6 & \textbf{98.7} \\
lamp & 1.67 & 3.42 & 1.80 & 1.17 & \textbf{0.90} & 1.47 & 2.76 & \textbf{0.795} & \textbf{0.795} & 62.7 & 35.0 & 84.0 & \textbf{89.1} & 87.1 & 62.9 & 73.8 & \textbf{93.5} & \textbf{93.5} \\
rifle & 0.19 & 0.42 & 0.09 & 1.07 & \textbf{0.05} & 0.39 & 0.55 & 0.060 & \textbf{0.057} & 86.2 & 70.0 & \textbf{97.3} & 93.5 & 96.2 & 86.1 & 81.1 & \textbf{96.9} & \textbf{96.9} \\
sofa & 0.30 & 0.80 & 0.35 & 0.13 & \textbf{0.11} & 0.31 & 0.16 & 0.208 & \textbf{0.037} & 85.9 & 55.2 & 92.8 & 92.5 & \textbf{93.5} & 85.2 & 89.3 & 98.3 & \textbf{98.4} \\
speaker & 1.01 & 1.99 & 0.68 & \textbf{0.14} & 0.27 & 0.38 & 0.17 & 0.065 & \textbf{0.044} & 74.7 & 47.4 & 84.3 & \textbf{90.2} & 90.1 & 78.1 & 89.4 & \textbf{97.3} & \textbf{97.3} \\
table & 0.44 & 1.57 & 0.56 & 0.17 & \textbf{0.13} & 0.31 & 0.30 & 0.107 & \textbf{0.046} & 84.9 & 55.7 & 92.4 & 93.4 & \textbf{93.7} & 87.2 & 88.6 & 96.5 & \textbf{97.6} \\
telephone & 0.13 & 0.39 & 0.08 & 0.08 & \textbf{0.06} & 0.19 & 0.11 & 0.043 & \textbf{0.010} & 94.8 & 81.8 & 98.1 & \textbf{98.8} & 98.3 & 88.9 & 96.5 & \textbf{99.6} & \textbf{99.6} \\
watercraft & 0.41 & 0.78 & 0.20 & 0.90 & \textbf{0.10} & 0.35 & 0.39 & 0.075 & \textbf{0.067} & 77.3 & 54.2 & 93.2 & 92.7 & \textbf{93.7} & 80.3 & 84.7 & \textbf{97.4} & 97.2 \\
\midrule
mean & 0.49 & 1.18 & 0.40 & 0.39 & \textbf{0.19} & 0.43 & 0.46 & 0.135 & \textbf{0.102} & 81.9 & 59.0 & 92.2 & 92.9 & \textbf{93.0} & 81.4 & 86.7 & 96.9 & \textbf{97.0} \\
\bottomrule
\end{tabular}
} %
\vspace{-0.5em}
\caption{\textbf{Per-category auto-encoding accuracy for objects in 3D-R$^2$N$^2$ test set of ShapeNet.} For each metric, left columns compare methods under encoder-decoder inference while right columns compare under decoder-only latent optimization. $*$: decoder-only latent optimization.}
\vspace{-0.5em}
\label{tab:quantitative_ae_full}
\end{table*}

\begin{table*}[h]
\centering
\resizebox{\linewidth}{!}{
\addtolength{\tabcolsep}{-3pt}  
\begin{tabular}{l||ccccc|cccc||ccccc|cccc}
\toprule
\multirow{2}{*}{Category} & \multicolumn{9}{c}{Chamfer~($\downarrow$)} & \multicolumn{9}{c}{F-Score~($\uparrow$, \%)}\\
    & Occ. & SIF & LDIF & IF. & Ours & Occ.* & IM.* & Local* & Ours* & Occ. & SIF & LDIF & IF. & Ours & Occ.* & IM.* & Local* & Ours* \\
\midrule
bed & 1.30 & 2.24 & 0.68 & \textbf{0.10} & 0.16 & 0.87 & 0.43 & 0.052 & \textbf{0.045} & 59.3 & 32.0 & 81.4 & \textbf{94.7} & 90.9 & 67.1 & 77.8 & 96.8 & \textbf{97.0} \\
birdhouse & 1.25 & 1.92 & 0.75 & 0.31 & \textbf{0.11} & 0.72 & 0.49 & \textbf{0.036} & \textbf{0.036} & 54.2 & 33.8 & 76.2 & 90.4 & \textbf{92.1} & 61.3 & 74.3 & 97.6 & \textbf{97.7} \\
bookshelf & 0.83 & 1.21 & 0.36 & 0.30 & \textbf{0.20} & 0.99 & 0.60 & 0.103 & \textbf{0.091} & 66.5 & 43.5 & 86.1 & \textbf{93.5} & 88.3 & 59.0 & 73.0 & \textbf{95.1} & 94.2 \\
camera & 1.17 & 1.91 & 0.83 & 0.27 & \textbf{0.16} & 0.45 & 0.58 & \textbf{0.047} & 0.050 & 57.3 & 37.4 & 77.7 & \textbf{95.0} & 94.0 & 70.2 & 75.9 & \textbf{98.6} & \textbf{98.6} \\
file & 0.41 & 0.71 & \textbf{0.29} & 0.35 & 0.30 & 0.38 & 0.25 & 0.054 & \textbf{0.041} & 86.0 & 65.8 & 93.0 & \textbf{95.7} & 94.4 & 84.3 & 90.0 & 97.6 & \textbf{97.7} \\
mailbox & 0.60 & 1.46 & 0.40 & 1.18 & \textbf{0.20} & 0.51 & 0.74 & \textbf{0.102} & \textbf{0.102} & 67.8 & 38.1 & 87.6 & 81.4 & \textbf{93.5} & 80.0 & 85.2 & \textbf{98.5} & \textbf{98.5} \\
piano & 1.07 & 1.81 & 0.78 & 0.34 & \textbf{0.08} & 0.91 & 0.71 & 0.034 & \textbf{0.030} & 61.4 & 39.8 & 82.2 & \textbf{96.7} & 94.8 & 62.2 & 77.3 & \textbf{98.3} & \textbf{98.3} \\
printer & 0.85 & 1.44 & 0.43 & \textbf{0.15} & \textbf{0.15} & 0.48 & 0.31 & \textbf{0.035} & \textbf{0.035} & 66.2 & 40.1 & 84.6 & \textbf{94.9} & 94.3 & 74.9 & 82.3 & 98.2 & \textbf{98.3} \\
stove & 0.49 & 1.04 & 0.30 & 0.55 & \textbf{0.22} & 0.37 & 0.25 & 0.107 & \textbf{0.040} & 77.3 & 52.9 & 89.2 & 91.3 & \textbf{93.5} & 78.6 & 87.4 & \textbf{97.7} & \textbf{97.7} \\
tower & 0.50 & 1.05 & 0.47 & 0.44 & \textbf{0.14} & 0.53 & 0.30 & \textbf{0.060} & 0.070 & 70.2 & 45.9 & 85.7 & 90.3 & \textbf{91.8} & 73.9 & 81.7 & \textbf{96.9} & \textbf{96.6} \\
\midrule
mean & 0.85 & 1.48 & 0.53 & 0.40 & \textbf{0.17} & 0.62 & 0.47 & 0.063 & \textbf{0.054} & 66.6 & 43.0 & 84.4 & 92.4 & \textbf{92.8} & 71.1 & 80.5 & \textbf{97.5} & \textbf{97.5} \\
\bottomrule
\end{tabular}
} %
\vspace{-0.5em}
\caption{\textbf{Per-category auto-encoding accuracy for objects in unseen categories of ShapeNet.} For each metric, left columns compare methods under encoder-decoder inference while right columns compare under decoder-only latent optimization. $*$: decoder-only latent optimization.}
\vspace{-1.5em}
\label{tab:unseen_ae_full}
\end{table*}

\subsection{Shape Completion User Study}

First, in~\Table{depth_comp}, we compare quantitative results of \MDIF and competing methods on shape completion from depth image.
In this comparison, we also include a \MDIF model (``Ours'') that uses encoder-decoder inference. This model has the same architecture as the \MDIF model in the point cloud completion experiment, and is retrained from scratch to take voxelized depth points (depth points voxelized into a $128^3$ grid) as input.
In terms of metrics, we additionally use Asymmetric Chamfer to measure the reconstruction accuracy in observed regions. It is computed as one-directional Chamfer L2 distance from depth points to reconstruction.

\begin{table}[H]
\centering
\resizebox{\linewidth}{!}{
\addtolength{\tabcolsep}{-3pt}  
\begin{tabular}{l||cc||cc}
\toprule
\multirow{2}{*}{Category} & \multicolumn{2}{c}{Point Cloud Completion} & \multicolumn{2}{c}{Voxel Super-Resolution}\\
    & IF-Net & Ours & IF-Net & Ours \\
\midrule
airplane & 2.37 / 89.7 & \textbf{0.08} / \textbf{93.3} & 1.51 / 78.3 & \textbf{1.02} / \textbf{80.7} \\
bench & 1.22 / 84.5 & \textbf{0.18} / \textbf{86.0} & 1.88 / 59.1 & \textbf{1.09} / \textbf{59.5} \\
cabinet & 1.65 / \textbf{87.1} & \textbf{0.84} / 83.8 & 0.65 / 60.6 & \textbf{0.60} / \textbf{60.8} \\
car & 1.96 / 79.4 & \textbf{0.19} / \textbf{80.9} & 0.40 / \textbf{75.8} & \textbf{0.30} / \textbf{75.8} \\
chair & 2.02 / \textbf{81.3} & \textbf{0.33} / 80.5 & 1.02 / 62.6 & \textbf{0.82} / \textbf{63.4} \\
display & 1.09 / 88.5 & \textbf{0.30} / \textbf{88.6} & 1.04 / 62.0 & \textbf{0.74} / \textbf{62.1} \\
lamp & 2.03 / 76.3 & \textbf{1.76} / \textbf{78.0} & 8.14 / 58.3 & \textbf{3.97} / \textbf{60.9} \\
rifle & 2.19 / 85.3 & \textbf{0.05} / \textbf{95.9} & 2.09 / 78.0 & \textbf{0.34} / \textbf{81.3} \\
sofa & 0.71 / \textbf{88.2} & \textbf{0.18} / 86.8 & 0.68 / 56.2 & \textbf{0.48} / \textbf{57.5} \\
speaker & 1.52 / \textbf{78.4} & \textbf{0.65} / 75.9 & 0.73 / 56.1 & \textbf{0.65} / \textbf{58.0} \\
table & 1.70 / 84.7 & \textbf{0.25} / \textbf{85.1} & 2.72 / 53.5 & \textbf{1.87} / \textbf{55.7} \\
telephone & 0.98 / 95.7 & \textbf{0.06} / \textbf{96.5} & 0.77 / 77.9 & \textbf{0.67} / \textbf{78.2} \\
watercraft & 1.51 / 87.2 & \textbf{0.14} / \textbf{88.4} & 2.05 / 71.7 & \textbf{0.69} / \textbf{73.6} \\
\midrule
mean & 1.61 / 85.0 & \textbf{0.39} / \textbf{86.1} & 1.82 / 65.4 & \textbf{1.02} / \textbf{66.7} \\
\bottomrule
\end{tabular}
} %
\vspace{-0.5em}
\caption{Per-category quantitative results (Chamfer L2 distance / F-Score) for point cloud completion and voxel super-resolution.}
\vspace{-0.5em}
\label{tab:pcc_vsr_full}
\end{table}

\begin{table*}[t]
\centering
\resizebox{0.85\linewidth}{!}{
\addtolength{\tabcolsep}{-2pt}  

\begin{tabular}{l||cccc||cccc||cccc}
\toprule
\multirow{2}{*}{Category} & \multicolumn{4}{c}{Chamfer~($\downarrow$)} &
\multicolumn{4}{c}{F-Score~($\uparrow$, \%)} & \multicolumn{4}{c}{Asym. Chamfer~($\downarrow$)}\\
    & OccNet & LDIF & Ours & Ours* & OccNet & LDIF & Ours & Ours* & OccNet & LDIF & Ours & Ours*\\
\midrule
airplane & 0.47 & \textbf{0.17} & 0.26 & 0.46 & 70.1 & 89.2 & \textbf{90.1} & 73.2 & 0.246 & 0.054 & 0.022 & \textbf{0.007} \\
bench & 0.70 & \textbf{0.39} & 0.45 & 0.96 & 64.9 & 81.9 & \textbf{82.5} & 56.9 & 0.281 & 0.108 & 0.049 & \textbf{0.012} \\
cabinet & 1.13 & 0.77 & \textbf{0.73} & 1.35 & 70.1 & \textbf{77.9} & 73.8 & 60.4 & 0.109 & 0.052 & 0.070 & \textbf{0.009} \\
car & 0.99 & 0.51 & \textbf{0.41} & 1.04 & 61.6 & 72.4 & \textbf{74.3} & 64.2 & 0.138 & 0.054 & 0.043 & \textbf{0.011} \\
chair & 2.34 & 1.02 & \textbf{0.91} & 1.42 & 50.2 & 69.6 & \textbf{72.5} & 67.0 & 0.785 & 0.270 & 0.053 & \textbf{0.012} \\
display & 0.95 & 0.62 & \textbf{0.56} & 1.69 & 62.8 & \textbf{80.0} & 76.7 & 55.4 & 0.312 & 0.217 & 0.056 & \textbf{0.007} \\
lamp & 9.91 & 2.15 & \textbf{1.26} & 3.26 & 44.1 & 66.4 & \textbf{70.5} & 54.6 & 10.80 & 1.429 & 0.160 & \textbf{0.110} \\
rifle & 0.49 & \textbf{0.14} & 0.31 & 0.62 & 66.4 & \textbf{92.3} & 91.5 & 75.9 & 0.246 & 0.048 & 0.022 & \textbf{0.005} \\
sofa & 1.08 & 0.83 & \textbf{0.70} & 1.19 & 61.2 & \textbf{71.7} & 71.4 & 62.1 & 0.155 & 0.074 & 0.059 & \textbf{0.007} \\
speaker & 3.50 & 1.48 & \textbf{1.45} & 3.73 & 52.4 & \textbf{67.3} & 64.6 & 49.8 & 0.280 & 0.115 & 0.077 & \textbf{0.020} \\
table & 2.49 & 1.14 & \textbf{0.94} & 1.11 & 66.7 & \textbf{78.0} & 77.8 & 61.5 & 0.784 & 0.339 & 0.065 & \textbf{0.015} \\
telephone & 0.35 & \textbf{0.19} & 0.21 & 1.05 & 86.1 & \textbf{92.0} & 89.4 & 55.9 & 0.089 & 0.046 & 0.046 & \textbf{0.002} \\
watercraft & 1.15 & 0.50 & \textbf{0.45} & 0.69 & 54.5 & 77.5 & \textbf{78.3} & 67.2 & 0.684 & 0.148 & 0.033 & \textbf{0.020} \\
\midrule
mean & 1.97 & 0.76 & \textbf{0.67} & 1.43 & 62.4 & \textbf{78.2} & 78.0 & 61.9 & 1.147 & 0.227 & 0.058 & \textbf{0.018}  \\
\bottomrule
\end{tabular}

\addtolength{\tabcolsep}{2pt}
} %
\caption{\textbf{Shape completion from depth image.} Quantitative comparisons on Chamfer distance, F-Score and Asymmetric Chamfer distance. 
``Ours*'' achieves the lowest error on the observed part, measured by the Asymmetric Chamfer distance. Its worse Chamfer and F-Score results are caused by the unobserved part. See our user study for more in-depth analysis. $*$: decoder-only latent optimization.}
\label{tab:depth_comp}
\end{table*}

When comparing under encoder-decoder inference (``OccNet'', ``LDIF'', ``Ours''), \MDIF is only slightly worse than LDIF in F-Score while performs the best in the other two metrics. This reveals that when using encoder-decoder inference, \MDIF can produce completion results similarly close to the groundtruth as LDIF.
Meanwhile, the large margin in Asymmetric Chamfer compared with OccNet and LDIF demonstrates the better capability of \MDIF to preserve details in observed regions, even under encoder-decoder inference.
For the \MDIF model that uses decoder-only latent optimization (``Ours*''), although it has worse performance in Chamfer distance and F-Score, it can reduce the error in Asymmetric Chamfer even much further.
This indicates that it performs much better on the observable parts and the source of error mostly comes from the unobserved parts.
As illustrated in the paper (\Fig{depth_comp}), although different from the groundtruth, the unobserved parts of its results still look plausible.

To prove our point, we conducted a user study to compare human subjective verdicts and F-Score. We recruited 88 participants who were at least 18 years old. All participants had no prior knowledge of this project. Each participant was given 32 pairs of examples, one from \MDIF (with decoder-only latent optimization) and one from LDIF~\cite{Genova_2020_CVPR}. Order of the examples is fully counterbalanced and randomized. Each example was shown in two different views: one observed (input view) and one unobserved. Participants were then asked to choose which example was the more plausible reconstruction given the input. If both examples looked similarly plausible, they were allowed to choose \textit{cannot decide}. 

Examples were chosen in this way. The worst results in F-Score were filtered, since both human and F-Score tend to agree on those cases. Then examples with unmatched input views were removed. We then randomly picked 32 examples from the rest. 

\begin{figure}[H]
\centering
\includegraphics[trim={0cm 8cm 18cm 0cm},clip,width=0.99\linewidth]{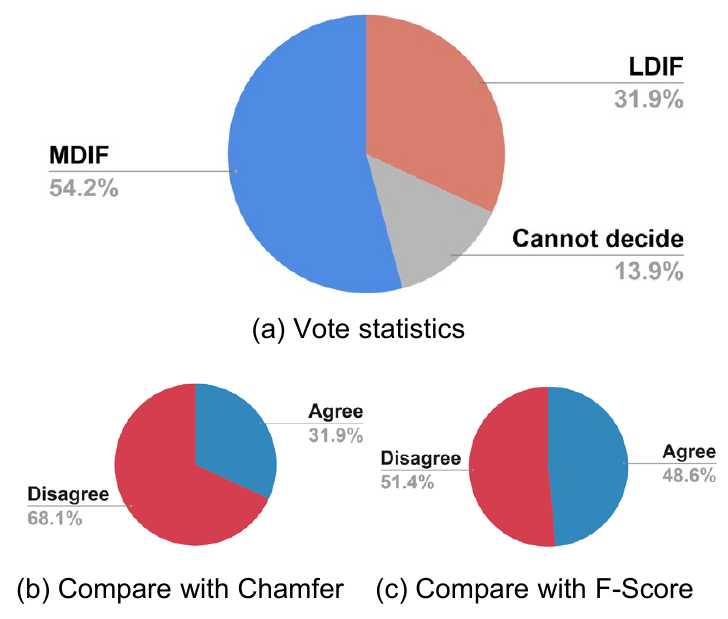}
\caption{\textbf{Summary of user study.} Participants were asked which reconstruction was more plausible. $54.2\%$ chose \MDIF while $13.9\%$ cannot decide between the results. Moreover, $68.1\%$ of the votes disagree with Chamfer L2 distance, and $51.4\%$ disagree with F-Score. 
Refer to \Figure{user_study_itemized_1}  to \Figure{user_study_itemized_4} for itemized results. }
\label{fig:user_study_summary}
\end{figure}

\begin{figure*}[h]
\centering
\includegraphics[trim={0cm 1cm 0cm 0.5cm},clip,width=1.0\textwidth]{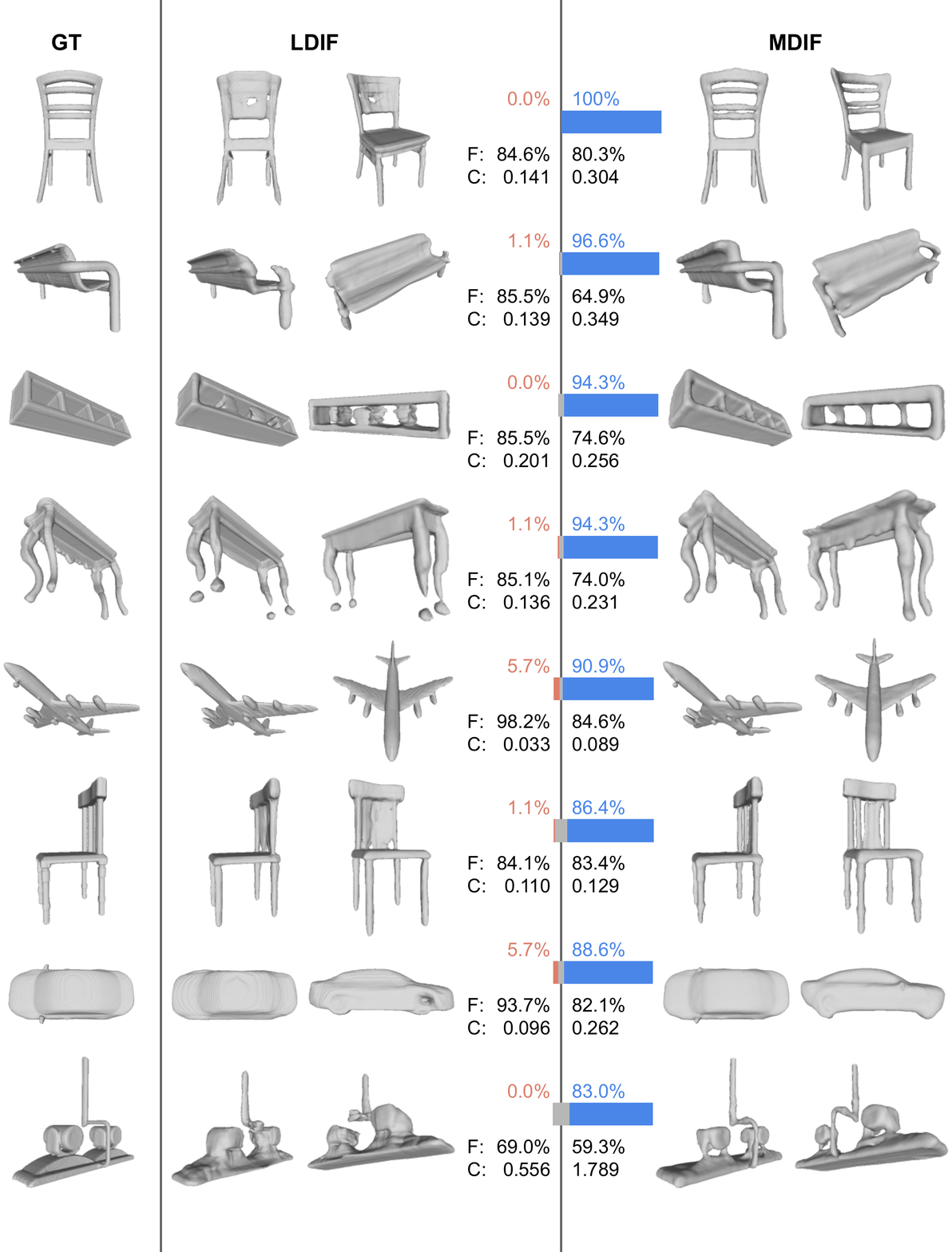}
\vspace{-1em}
\caption{\textbf{Itemized user study results.} For each example, we show the groundtruth mesh under input view, and the reconstruction results under two views: one observed view same as input and one unobserved view. The bar chart shows the percentages of votes. Red: prefer LDIF; Blue: prefer MDIF; Gray: Cannot decide; F: F-Score; C: Chamfer L2 distance.}
\vspace{-1em}
\label{fig:user_study_itemized_1}
\end{figure*}

\begin{figure*}[h]
\centering
\includegraphics[trim={0cm 1cm 0cm 0.5cm},clip,width=1.0\textwidth]{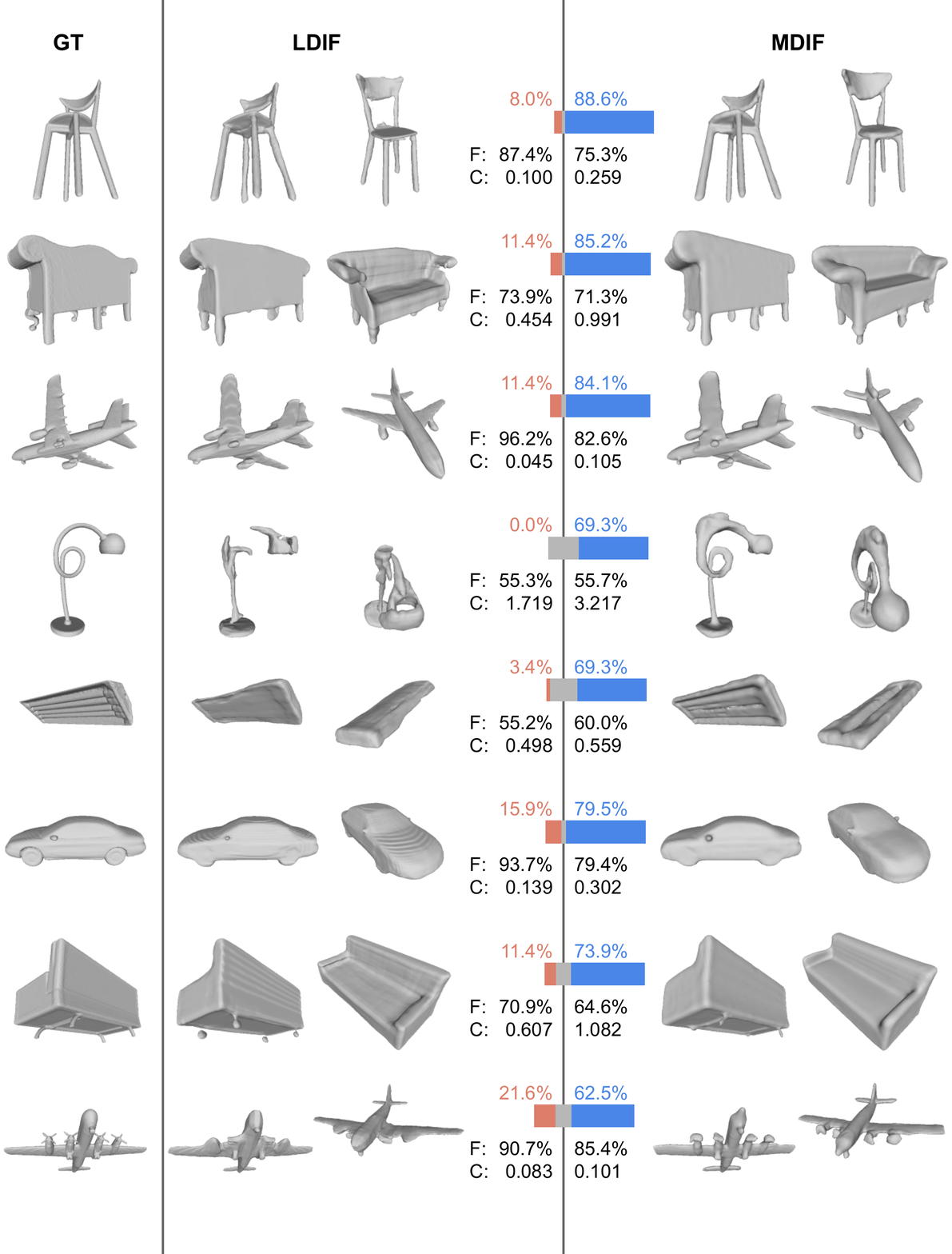}
\vspace{-1em}
\caption{\textbf{Itemized user study results.} For each example, we show the groundtruth mesh under input view, and the reconstruction results under two views: one observed view same as input and one unobserved view. The bar chart shows the percentages of votes. Red: prefer LDIF; Blue: prefer MDIF; Gray: Cannot decide; F: F-Score; C: Chamfer L2 distance.}
\vspace{-1em}
\label{fig:user_study_itemized_2}
\end{figure*}

\begin{figure*}[h]
\centering
\includegraphics[trim={0cm 1cm 0cm 0.5cm},clip,width=1.0\textwidth]{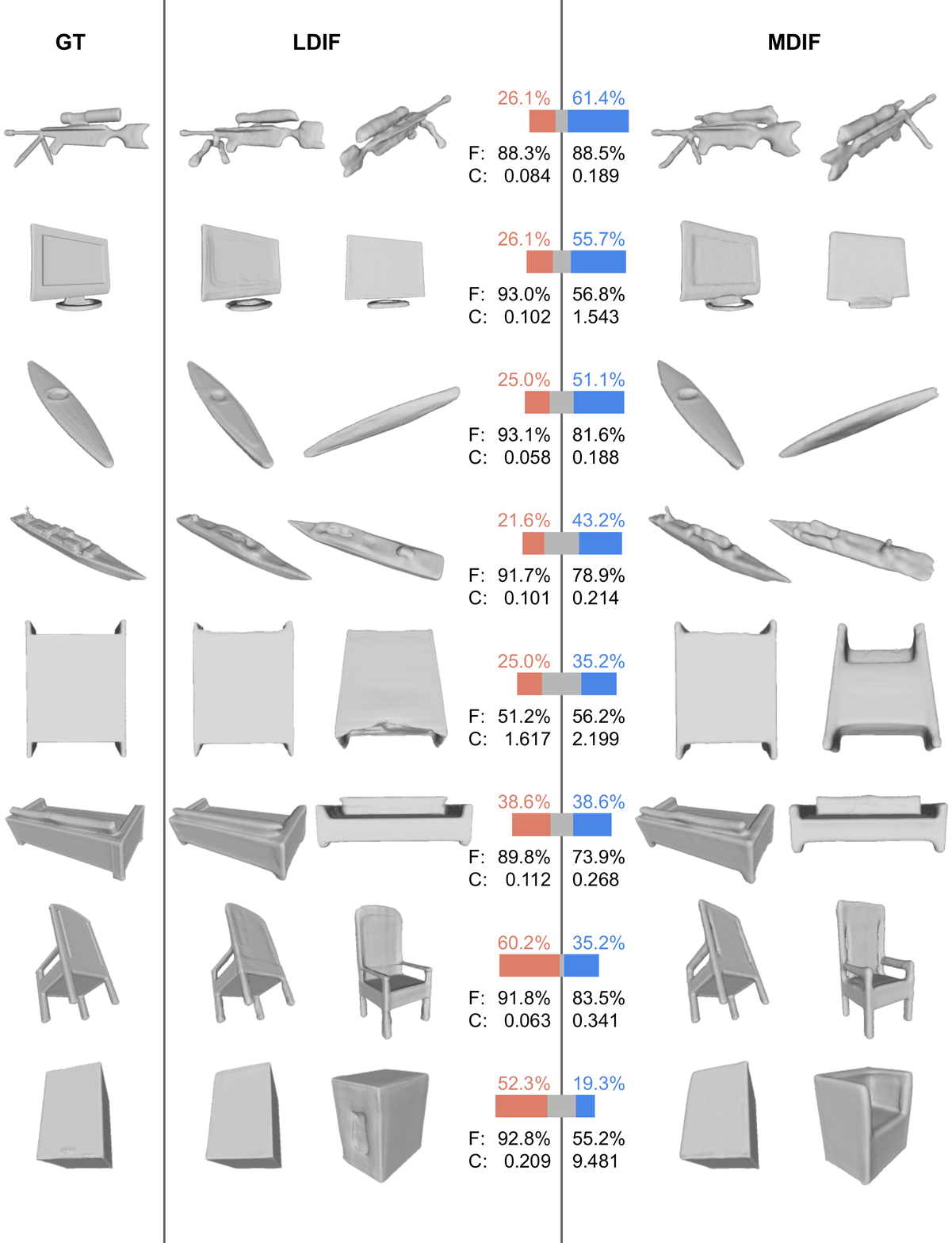}
\vspace{-1em}
\caption{\textbf{Itemized user study results.} For each example, we show the groundtruth mesh under input view, and the reconstruction results under two views: one observed view same as input and one unobserved view. The bar chart shows the percentages of votes. Red: prefer LDIF; Blue: prefer MDIF; Gray: Cannot decide; F: F-Score; C: Chamfer L2 distance.}
\vspace{-1em}
\label{fig:user_study_itemized_3}
\end{figure*}

\begin{figure*}[h]
\centering
\includegraphics[trim={0cm 1cm 0cm 0.5cm},clip,width=1.0\textwidth]{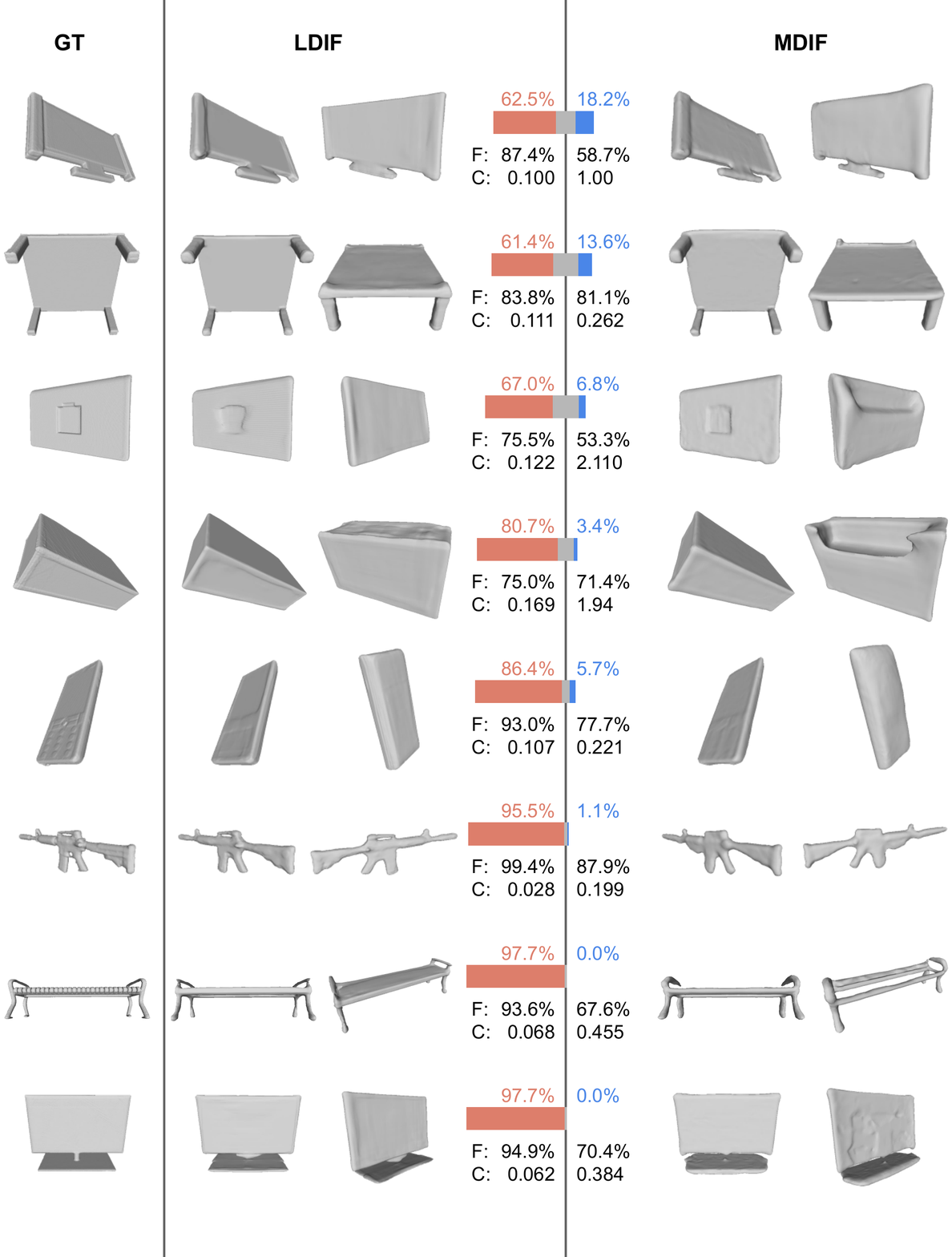}
\vspace{-1em}
\caption{\textbf{Itemized user study results.} For each example, we show the groundtruth mesh under input view, and the reconstruction results under two views: one observed view same as input and one unobserved view. The bar chart shows the percentages of votes. Red: prefer LDIF; Blue: prefer MDIF; Gray: Cannot decide; F: F-Score; C: Chamfer L2 distance.}
\vspace{-1em}
\label{fig:user_study_itemized_4}
\end{figure*}

The results of user study are summarized in~\Fig{user_study_summary}. In contrast to F-Score, $54.2\%$ of the participants chose in favor of \MDIF results, whilst $31.9\%$ thought LDIF results were better. In addition, $13.9\%$ could not decide between \MDIF and LDIF. 
Moreover, when compared with the quantitative metrics, $68.1\%$ disagree with Chamfer L2 distance, and $51.4\%$ disagree with F-Score. 
All the 32 examples and itemized results are shown in~\Figure{user_study_itemized_1},~\Figure{user_study_itemized_2},~\Figure{user_study_itemized_3} and~\Figure{user_study_itemized_4}.

The conclusion of this user study aligns with previous work~\cite{tatarchenko2019single}, where Chamfer distance has been argued as not suitable for evaluating completion tasks due to its sensitivity to outliers. Moreover, this study also shows that, although more robust, F-Score only tells us how different the reconstruction of the unobserved part is from the groundtruth, but not how \textit{plausible} it is, which is what humans ultimately care about.

\clearpage

